\documentclass[final,3p,times]{elsarticle}

\usepackage{amssymb}
\usepackage[linesnumbered,ruled,lined]{algorithm2e}
\usepackage{amsmath}
\usepackage{subcaption}
\usepackage{siunitx}
\usepackage{multirow}
\usepackage{hyperref}
\usepackage{booktabs}
\usepackage{multirow}
\usepackage{natbib}

\usepackage{xcolor}

\SetKwComment{Comment}{$\#$ }{}
\SetCommentSty{mycommfont}

\journal{Knowledge-Based Systems}

\begin{document}

\begin{frontmatter}



\title{Direct Interval Propagation Methods using Neural-Network Surrogates for Uncertainty Quantification in Physical Systems Surrogate Model}


\author[inst1,inst5]{Ghifari Adam Faza}
\author[inst2]{Jolan Wauters}
\author[inst4]{Fabio Cuzzolin}
\author[inst3]{Hans Hallez}
\author[inst1,inst5]{David Moens}

\affiliation[inst1]{organization={LMSD, Department of Mechanical Engineering, KU Leuven},
            city={Heverlee},
            postcode={3001},
            country={Belgium}}

\affiliation[inst2]{organization={RAM, Department of Mechanical Engineering, KU Leuven},
            city={Brugge},
            postcode={8200},
            country={Belgium}}

\affiliation[inst3]{organization={DistriNet, Department of Computer Science, KU Leuven},
            city={Brugge},
            postcode={8200},
            country={Belgium}}

\affiliation[inst4]{organization={Visual Artificial Intelligence Laboratory, Oxford Brookes University},
            city={Oxford},
            country={UK}}
            
\affiliation[inst5]{organization={FlandersMake@KU Leuven},
            country={Belgium}}

\begin{abstract}
In engineering applications, uncertainty propagation refers to the systematic characterisation of a system output under uncertain inputs. Specifically for interval uncertainty, the objective is to determine the lower and upper bounds of the output, given interval-valued inputs. Such uncertainty propagation plays a crucial role in engineering tasks, including robust design optimisation and reliability analysis, where accurate characterisation of uncertainty is essential for safe and reliable decision-making. However, standard interval propagation requires solving optimisation problems that can be computationally expensive, particularly when dealing with complex physical systems. To address this challenge, surrogate models have been developed to enable efficient interval propagation. Although surrogate models are computationally more efficient, standard surrogate-based approaches typically only replace the evaluator function within the optimisation loop, which still requires a large number of inference calls. Therefore, we propose to directly estimate the output interval by reframing the problem as an interval-valued regression task. In this work, we present a comprehensive study of strategies for direct interval propagation using NN-based surrogate models, including standard multilayer perceptrons (MLPs) and deep operator networks (DeepONet). We investigate and compare three distinct approaches: (i) naive interval propagation through standard architectures, (ii) bound propagation techniques such as Interval Bound Propagation (IBP) and CROWN, and (iii) interval neural networks (INNs) with interval weights. Our results demonstrate that these methods are significantly more efficient compared to traditional optimisation-based interval propagation and are able to provide accurate interval estimates. We also discuss the limitations and open challenges associated with implementing interval-based propagation in practice.
\end{abstract}


\begin{highlights}
\item We provide a comprehensive evaluation of direct interval uncertainty propagation strategies in neural network surrogate models, benchmarking them against optimisation-based propagation. Specifically, we compare naive interval propagation, bound propagation, and interval neural networks with interval-valued weights, highlighting their respective strengths and limitations.
\item We introduce a modified DeepONet architecture capable of directly handling interval-valued variables. This extension enables operator-learning-based surrogates to propagate interval uncertainties across the input and output spaces, making them applicable to full-field problems commonly encountered in computational mechanics.  
\item We propose a data augmentation strategy that allows direct interval propagation models to be trained when only pointwise datasets are available. This approach alleviates the scarcity of interval-valued training data, which are often expensive or impractical to obtain, and broadens the applicability of interval UQ in real-world engineering scenarios. 
\item We analyse the difficulties that arise when training operator-learning models with augmented interval datasets and propose a practical workaround to mitigate these issues.
\end{highlights}


\end{frontmatter}



\section{Introduction}
\label{sec: intro}
In engineering applications, uncertainty propagation refers to the systematic characterisation of how input uncertainties affect system outputs. Following \cite{Abdi2023, Dai2024}, we distinguish uncertainty propagation, which transforms known input uncertainties through a model, from uncertainty prediction, which estimates output confidence or variability without explicit input uncertainty. Classical approaches in propagation, such as Monte Carlo (MC) and quasi-Monte Carlo (qMC) simulations~\cite{Dick2013, Lemieux2009}, provide probabilistic estimates of output variability by repeatedly sampling the input space. While these methods are conceptually straightforward, they are computationally expensive, particularly for high-fidelity simulations such as finite element or computational fluid dynamics models, where a single run can require hours on high-performance computing resources. To alleviate this burden, surrogate models such as Gaussian processes~\cite{Rasmussen2005-at, Bilionis2012, Zuhal2023}, polynomial chaos expansion~\cite{Ghanem1990, Sudret2008}, and neural networks~\cite{tripathy2018deep, sun2019review} are increasingly employed to approximate the input-output mapping, enabling efficient uncertainty propagation without the need for repeated costly simulations.

This work focuses on interval uncertainty propagation, where uncertainty is represented by lower and upper bounds rather than full distributions. Interval methods are advantageous when data are scarce or probabilistic assumptions are unreliable. For instance, in structural design, material properties may vary due to manufacturing, but only a few tests may be available; representing these properties as intervals provides a robust description of uncertainty. Our goal is to propagate these input intervals (e.g., material properties) through the model to predict output intervals (e.g., stresses or displacements), which can then inform safety assessment, reliability analysis, or design optimisation.

Standard interval propagation methods, described in Section~\ref{sec: classic_IP}, require solving two optimisation problems per point to determine the interval bounds. Surrogate models can reduce the cost of evaluating the underlying simulator, but they still function as emulators that must be queried repeatedly, which can remain expensive. This inefficiency is amplified in complex problems requiring full-field solutions, where intervals must be propagated across every mesh point. Directly designing the surrogate to output interval estimates can bypass repeated evaluations and substantially improve efficiency.

In this work, we investigate strategies for directly propagating input uncertainties through to the output space using neural network-based surrogate models. Specifically, we compare three distinct approaches: (i) naive interval propagation through a standard neural network architecture~\cite{Yang2019}, (ii) bound propagation using the IBP~\cite{gowal2019} and CROWN~\cite{zhang2018} algorithm, and (iii) interval neural networks with interval-valued weights~\cite{Oala2021, Betancourt2022, Tretiak2023}, and compare them with the optimisation-based propagation baseline. 

To assess their performance, we consider two scenarios. In the first, we assume full access to an interval-valued dataset (i.e., paired interval inputs and outputs). Since such datasets are rare and costly to obtain, we also evaluate the methods when only pointwise data are available for training and evaluate the model on actual interval data. For each scenario, we conduct both a simple one-dimensional regression test, serving as a proof of concept, and some more challenging full-field partial differential equations (PDEs) surrogate modelling tasks, which is particularly well suited to operator-learning architectures such as deep operator networks (DeepONet)~\cite{Lu2021}. To summarise, the following are the main contributions of this paper:
\begin{enumerate}
    \item We provide a comprehensive evaluation of direct interval uncertainty propagation strategies in neural network surrogate models, benchmarking them against optimisation-based propagation. Specifically, we compare naive interval propagation, bound propagation, and interval neural networks with interval-valued weights, highlighting their respective strengths and limitations.
    \item We introduce a modified DeepONet architecture capable of directly handling interval-valued variables. This extension enables operator-learning-based surrogates to propagate interval uncertainties across the input and output spaces, making them applicable to full-field problems commonly encountered in computational mechanics.  
    \item We propose a data augmentation strategy that allows direct interval propagation models to be trained when only pointwise datasets are available. This approach alleviates the scarcity of interval-valued training data, which are often expensive or impractical to obtain, and broadens the applicability of interval UQ in real-world engineering scenarios. 
    \item We analyse the difficulties that arise when training operator-learning models with augmented interval datasets and propose a practical workaround to mitigate these issues.
\end{enumerate}

The remainder of this paper is organised as follows: Section~\ref{sec: classic_IP} introduces the fundamentals of interval propagation methods and highlights their key challenges. Section~\ref{sec: direct_IP} presents the formulation of three direct interval propagation approaches. The experimental setup is described in Section~\ref{sec: experiments}, with Section~\ref{sec: exp_ideal} focusing on experiments under ideal assumptions and Section~\ref{sec: exp_aug} addressing the case where only pointwise training data are available. Finally, Section~\ref{sec: discussion} concludes the paper with a discussion of the results and the future works.

\section{Classical interval propagation methods}
\label{sec: classic_IP}

Consider a generic deterministic model $\mathcal{M}$ used in an engineering system, which takes an input parameter vector $\boldsymbol{x}$ and produces an output quantity of interest (QoI), denoted as $\boldsymbol{y}$. The model $\mathcal{M}$ may represent, for instance, a finite element model, a boundary element method, a deterministic analytical function, or any other deterministic modelling approach. This mapping can be formally expressed as:
\begin{equation}
    \mathcal{M} : \boldsymbol{x} \mapsto \boldsymbol{y}.
\end{equation}
When the input parameters are subject to interval uncertainty, they are represented by an interval vector
    $\boldsymbol{x}^\dagger = [\boldsymbol{x}_L, \boldsymbol{x}_U] = \left[ x_1^{\text{int}}, x_2^{\text{int}}, \ldots, x_r^{\text{int}} \right]^T,$
where $ x_i^{\text{int}} = [x_{L_i}, x_{U_i}], \ i=1,2,\ldots,r $ denotes the interval for the $ i $-th variable, and $ \boldsymbol{x}_L $ and $ \boldsymbol{x}_U $ are the vectors of lower and upper bounds for all input variables.

Given this interval-valued input, the corresponding bounds on the QoI can be computed by solving the following optimisation problems:
\begin{equation}
    y_L = \min_{\boldsymbol{x}_L \leq \boldsymbol{x} \leq \boldsymbol{x}_U} \mathcal{M}(\boldsymbol{x}), \quad 
    y_U = \max_{\boldsymbol{x}_L \leq \boldsymbol{x} \leq \boldsymbol{x}_U} \mathcal{M}(\boldsymbol{x}).
    \label{eq: interval_prop}
\end{equation}
These expressions yield the minimum and maximum values of the system response over all possible evaluations within the input interval vector $\boldsymbol{x}^\dagger$, assuming that each uncertain parameter varies independently within its respective bounds. It is important to note that, due to the potential non-monotonic behaviour of the model  $\mathcal{M}$, these optimisation problems generally cannot be solved analytically and must be evaluated numerically. Consequently, obtaining tight output bounds requires exploring a high-dimensional input space, which makes interval propagation computationally expensive, especially when the number of uncertain parameters is large or the model is highly nonlinear. This motivates the use of surrogate models to enable more efficient interval propagation.

To mitigate the high computational cost of interval propagation on complex models, surrogate modelling techniques are commonly employed as an emulator. A surrogate model $\hat{f}$ serves as a computationally efficient approximation of the original system, trained to replicate its input-output behaviour with significantly reduced evaluation time. This allows for faster uncertainty propagation while maintaining a reasonable level of accuracy. Among the various surrogate modelling approaches, neural network-based models, such as standard multilayer perceptrons (MLPs)~\cite{tripathy2018deep, White2019, Zhang2021} and operator learning~\cite{Lu2021, Kovachki2021} architectures, have shown considerable promise due to their expressiveness and scalability. Therefore, the optimisation problems in equation \ref{eq: interval_prop} can be reformulated as:
\begin{equation}
    y_L = \min_{\boldsymbol{x}_L \leq \boldsymbol{x} \leq \boldsymbol{x}_U} \hat{f}(\boldsymbol{x}), \quad 
    y_U = \max_{\boldsymbol{x}_L \leq \boldsymbol{x} \leq \boldsymbol{x}_U} \hat{f}(\boldsymbol{x}).
    \label{eq: interval_prop_surr}
\end{equation}

As alternatives to classical optimisation-based approaches for non-monotonic interval propagation, Cicirello et al. \cite{Cicirello2022} reformulate the problem using Bayesian optimisation. Although the underlying objective remains unchanged, this reformulation offers advantages such as faster convergence, reduced evaluations of the expensive model, and improved accuracy compared to optimisation over a full surrogate model. In a different approach, Callens et al. \cite{Callens2022} propose a multilevel quasi-Monte Carlo method for interval analysis, where intervals are modelled using Cauchy random variables to enable probabilistic sampling. This method demonstrates a significant reduction in computational cost. Given that the test cases in this study are inexpensive to evaluate, we adopt direct optimisation for interval propagation as the baseline for the benchmarking study.

Nonetheless, the double optimisation approach with surrogate models still requires numerous inference calls to the surrogate $\hat{f}$. This problem limits the scalability of the method, particularly for high-dimensional input spaces or when a full-field solution is required. In such cases, the optimisation must be performed for every discretisation point in the computational mesh, leading to a significant increase in computational cost. Denoting that $N_{\text{mesh}}$ the number of discretisation points, $N_{\text{samp}}$ the number of interval-valued instances, $N_{\text{iter}}$ the number of optimisation iterations, and $C_{\text{pred}}$ the computational cost of a single surrogate inference, the total computational cost is
$
    C_{\text{total}} = 2 \times N_{\text{mesh}} \times N_{\text{samp}} \times N_{\text{iter}} \times C_{\text{pred}}.
$
To address this issue, we propose to directly estimate the output interval by reframing the problem as an interval-valued regression task. This approach allows us to learn a direct mapping from interval-valued inputs to interval-valued outputs, eliminating the need for repeated surrogate evaluations during the optimisation process.

\section{Direct interval propagation}
\label{sec: direct_IP}

Direct Interval Propagation (DIP) is a term we use to describe the direct estimation of output intervals from input intervals using machine learning (ML) techniques. Unlike conventional approaches, where ML models serve as surrogates to emulate expensive optimisation procedures, DIP aims to learn a direct mapping from interval-valued inputs to interval-valued outputs:
\begin{equation}
    \hat{f} : [\boldsymbol{x}_L, \boldsymbol{x}_U] \mapsto [\boldsymbol{y}_L, \boldsymbol{y}_U].
\end{equation}
This task falls under the broader category of regression with interval-valued variables. In the context of linear models, several methods have already been proposed in the literature~\cite{Billard2000, LimaNeto2010}. For more complex engineering applications, Faza et al.~\cite{Faza2024} introduced an interval surrogate model based on interval proper orthogonal decomposition and polynomial chaos expansion. Similarly, Liu et al.~\cite{Liu2026} employ reduced-order models with interval-valued variables.

In this work, however, we focus exclusively on direct interval propagation using neural network-based architectures, namely standard multilayer perceptrons (MLPs) as a proof of concept and deep operator networks (DeepONet)~\cite{Lu2021} for full-field problems. In this section, we present three direct interval propagation strategies: (i) naive direct interval propagation, (ii) bound propagation methods (including IBP and CROWN), and (iii) interval neural networks. We also provide schematic illustrations of how these approaches can be applied within the DeepONet framework. Finally, we discuss the conceptual connection between interval propagation and interval estimation.

\subsection{Naive direct interval propagation}
\label{sec: naive_dip}

\subsubsection{Standard implementation}

\begin{figure}[ht]
    \begin{center}
    \includegraphics[width=.3\textwidth]{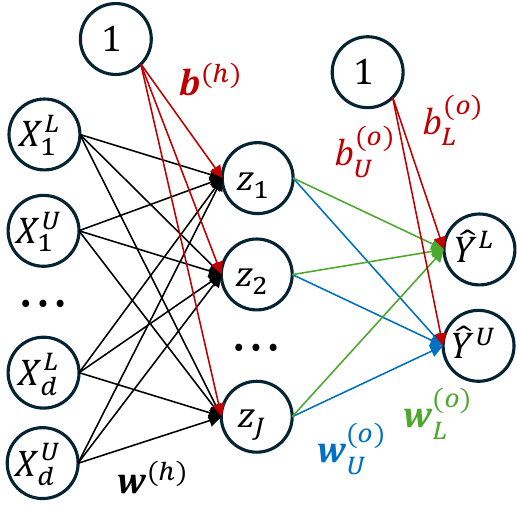}
    \end{center}
\caption{Regularised artificial neural network (RANN) architecture for naive direct interval propagation.}
\label{fig: naive_dip}
\end{figure}

The term naive DIP refers to any neural network architecture that retains a standard structure but extends the input and output layers to explicitly represent the lower bounds (LB) and upper bounds (UB) of the input and output variables. These models do not incorporate interval arithmetic or interval-valued weights within the network. Several modelling paradigms exist for naive DIP. Some approaches represent intervals using the centre and range formulation~\cite{Roque2007,Maia2008}, while others directly model the lower and upper bounds~\cite{Maia2011, Yang2019}. To prevent crossing between the predicted bounds, Roque et al.~\cite{Roque2007} constrain the connection weights in the half-range regression to be non-negative. In contrast, Yang et al.~\cite{Yang2019} introduce a soft regularisation term in the loss function to penalise bound violations. The general architecture of the Naive approach, following the regularised artificial neural network (RANN)~\cite{Yang2019}, is illustrated in Figure \ref{fig: naive_dip}.

In a simple regression setting, the interval-valued data $(\boldsymbol{X}^\dagger, Y^\dagger)$ consist of $d$-dimensional predictors and one response. All of the data is in lower-upper form $\{[X_{1,L}, X_{1,U}], \ldots, [X_{d,L}, X_{d,U}] \}$ and $\{ Y_L, Y_U \}$. Assume that the model has 3 layers consisting of $2d$ input nodes, $J$ nodes in the hidden layers, and 2 output nodes. The model output is written as:
\begin{equation}
    \hat{\boldsymbol{y}}_L = \sigma^{(o)} \left( \sum_{j=1}^J z_j w_{j,L}^{(o)} + b_L^{(o)}\right) \text{ and } \hat{\boldsymbol{y}}_U = \sigma^{(o)} \left( \sum_{j=1}^J z_j w_{j,U}^{(o)} + b_U^{(o)}\right),
\end{equation}
where $w_{j,L}^{(o)}$ and $w_{j,U}^{(o)}$ are the weight between the hidden layer nodes and the lower and upper output nodes, respectively. The variable $b_L^{(o)}$ and $b_U^{(o)}$ are the bias terms of the output. The activation function of the outputs is represented as $\sigma^{(o)}$ and variable $z_j$ denotes the output of the $j$th hidden layer node, defined as:
\begin{equation}
    z_j = \sigma^{(h)} \left( \sum_{k=1}^{2d} X_k w_{k,j}^{(h)} + b_j^{(h)}\right).
\end{equation}
The superscript $(h)$ denotes that the activation function $\sigma^{(h)}$, weight $w_{k,j}^{(h)}$, and bias $b_j^{(h)}$ belong to the hidden layer nodes.

To train such neural network model, we typically minimise the mean squared error (MSE) loss function through the gradient descent method. However, in interval-valued data, we have to be consistent with the mathematical definition of the interval, such that $\hat{Y}_U \geq \hat{Y}_L$. Therefore, Yang et al.~\cite{Yang2019} introduce a non-crossing regulariser to meet the requirements. Given $N$ training data, the loss function is formulated as:
\begin{equation}
    \mathcal{L} = \frac{1}{N} \sum_{i=1}^N (y_{i,L} - \hat{y}_{i,L})^2 + \frac{1}{N} \sum_{i=1}^N (y_{i,U} - \hat{y}_{i,U})^2 + \frac{\lambda}{N} \sum_{i=1}^N \left( \max\{0, \hat{y}_{i,L} - \hat{y}_{i,U} \} \right)^2,
\end{equation}
where $\lambda \geq 0$ is the parameter for controlling the regularisation strength. When $\lambda$ is 0, the regularisation terms collapse to zero and the regularised model becomes identical with the model proposed by Maia et al.~\cite{Maia2011}. When $\lambda$ is large, then the regularisation term will dominate the loss function and the model will approximate the behaviour of the one proposed by Roque~\cite{Roque2007}. When $\lambda$ is moderate, then the model provide a flexible choice of a more realistic model that prevents most of the interval crossing with small negative effects to the prediction accuracy.

\subsubsection{Naive direct interval propagation DeepONet implementation}
This framework can also be applied to operator learning, a setting where modelling problems in engineering and the physical sciences often rely on computationally expensive simulations or resource-intensive experiments for solving complex systems of partial differential equations (PDEs). These costs severely constrain the number of feasible simulations or measurements, limiting downstream tasks such as reliability analysis, design optimisation, and inverse problem solving. To overcome these bottlenecks, faster and more scalable alternatives are essential. While advances in hardware acceleration have improved the efficiency of numerical PDE solvers~\cite{Huthwaite2014, Bernardini2021}, surrogate modelling techniques, such as the ``\emph{learning to simulate}'' with Graph Neural Networks (GNN)~\cite{sanchezgonzalez2020}, Deep Operator Networks (DeepONet)~\cite{Lu2021}, Neural Operators (NO)~\cite{Kovachki2021}, and the Universal Physics Transformer (UPT)~\cite{alkin2025}, are rapidly gaining popularity as promising substitutes. 

Let $D \subset \mathbb{R}^d$ be a bounded, open domain, and let $
\mathcal{U} = \mathcal{U}(D; \mathbb{R}^{d_u}), \ 
G = G(D; \mathbb{R}^{d_g})
$
denote separable Banach spaces of functions defined on $D$, taking values in $\mathbb{R}^{d_u}$ and $\mathbb{R}^{d_g}$, respectively. We consider a (typically nonlinear) operator
$
\mathcal{G}^\dagger : \mathcal{U} \to G
$
which maps an input function $u(\cdot) \in \mathcal{U}$ to an output function $g(\cdot) = \mathcal{G}^\dagger(u) \in G$. Such operator mappings arise naturally in forward models of PDEs, where $u$ represents coefficients, forcing terms, or boundary conditions, and $g$ denotes the corresponding solution field. Suppose that we have a set of $N$ observations $\{ u^{(i)}, g^{(i)} \}^N_{i=1}$, the goal is to build an approximation of $\mathcal{G}^{\star}$ by constructing:
\begin{equation}
    \mathcal{G}_{\boldsymbol{\theta}} : \mathcal{U} \rightarrow G, \quad \boldsymbol{\theta} \in \mathbb{R}^p,
\end{equation}
then optimising paratemer $\boldsymbol{\theta} \in \mathbb{R}^p$, where $p$ is the dimensionality of the parameter, so that $\mathcal{G}_{\boldsymbol{\theta}^\star} \approx \mathcal{G}^\star$.

In this paper, we employ DeepONet~\cite{Lu2021} as our operator learning model due to its simple architecture. Based on the universal approximation theorem of operators~\cite{chen1995}, the DeepONet architecture is comprised of two sub-networks: a trunk net, and a branch net. The trunk net is responsible for extracting the latent representation of the space-time location of the PDE $\boldsymbol{y} = [t,\boldsymbol{x}]$, where $t$ is time, and $x$ and $y$ are spatial location. Meanwhile, the branch net is responsible for handling the input functions $\boldsymbol{u} = [\boldsymbol{u}(\boldsymbol{x}_1), \ldots, \boldsymbol{u}(\boldsymbol{x}_m)]$ in the form of discretised $m$-points. Assuming that the latent representations of both trunk net $\boldsymbol{\tau} \in \mathbb{R}^q$ and branch net $\boldsymbol{\beta} \in \mathbb{R}^q$ are $q$-dimensional, where $\boldsymbol{\tau} = [\tau_1, \ldots, \tau_q]$ and $\boldsymbol{\beta} = [\beta_1, \ldots, \beta_q]$. The solution operators $\mathcal{G}$ of a function $\boldsymbol{u}$ evaluated at $\boldsymbol{x}$, is expressed as:
\begin{equation}
    \mathcal{G}(\boldsymbol{u})(\boldsymbol{x}) \approx \sum_{i=1}^q \beta_i(\boldsymbol{u}) \cdot \tau_i (\boldsymbol{x}).
\end{equation}

\begin{figure}[ht]
    \begin{center}
        \begin{subfigure}{.44\textwidth}
            \centering
            \includegraphics[width=.95\textwidth]{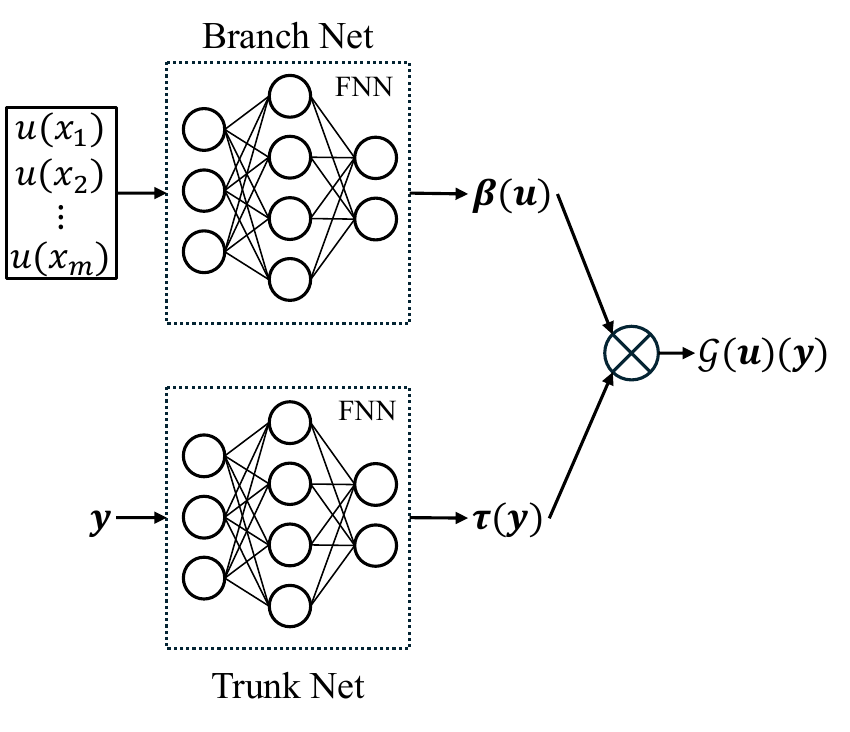}
            \caption{}
            \label{fig: deeponet}
        \end{subfigure}
        \begin{subfigure}{.54\textwidth}
            \centering
            \includegraphics[width=.95\textwidth]{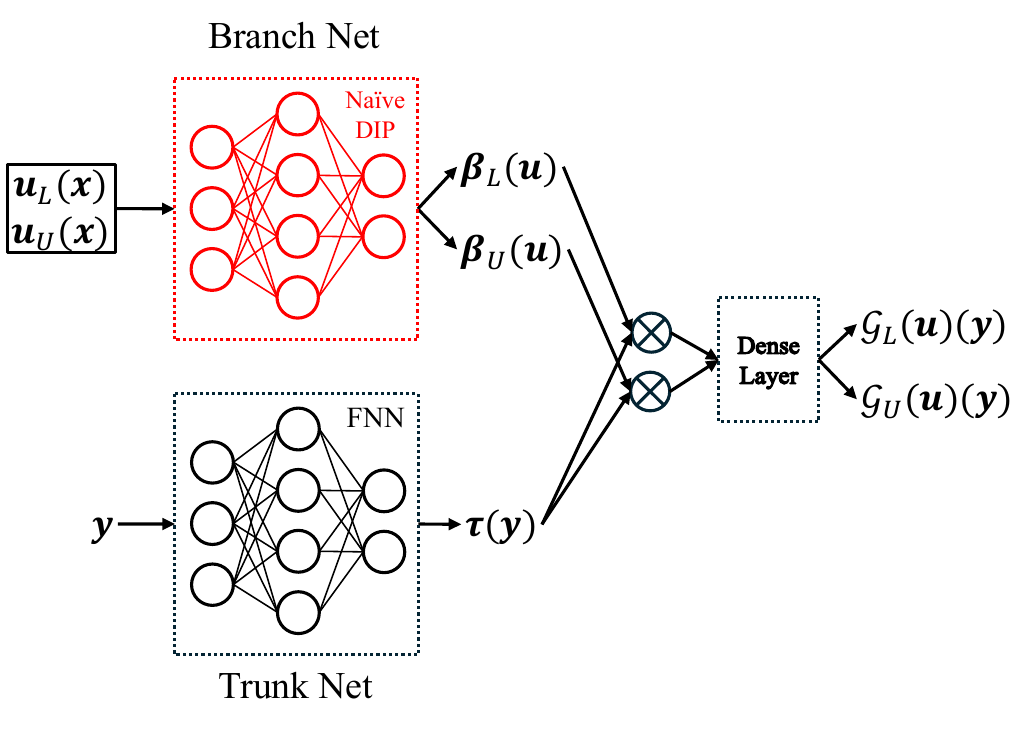}
            \caption{}
            \label{fig: naivedeeponet}
        \end{subfigure}
    \end{center}
\vspace{-0.7cm}
\caption{Architecture illustration for (a) Standard DeepONet architecture. (b) Modified DeepONet architecture with interval-valued branch net. The standard architecture use standard feedforward neural network (FNN) on both branches, meanwhile the modified network replance FNN with the Naive interval architecture.}
\end{figure}

In contrast to other neural network-based PDE solvers that rely on specialized architectural components, such as graph layers in GNNs~\cite{sanchezgonzalez2020}, integral kernel layers in Neural Operators (NOs)~\cite{Kovachki2021}, or transformer layers in Universal Physics Transformers (UPTs)~\cite{alkin2025}, the DeepONet architecture employs a more straightforward design. Its trunk and branch networks can be implemented using standard fully connected neural networks (FNN), making the model structurally simpler and easier to manipulate. This simplicity facilitates rapid prototyping and allows for seamless adaptation to interval-valued data. The DeepONet architecture illustration is given in Figure \ref{fig: deeponet}.

In this study, we assume that only the input function $\boldsymbol{u}$ is the interval-valued input, simply because in a real-world setting, the space-time location of the PDE is usually known. Therefore, we only replace the branch net with the naive DIP. The illustration of the modified architecture is given in Figure \ref{fig: naivedeeponet}. The latent representation of the branch net is now $2q$-dimensonal, given by $\boldsymbol{\beta^\dagger} = [\beta_{1,L}, \beta_{1,U}, \ldots, \beta_{q,L}, \beta_{q,U}]$. Since we now have $\boldsymbol{\tau} \in \mathbb{R}^q$ and $\boldsymbol{\beta} \in \mathbb{R}^{2q}$, we combine the two latent representations by using two multiplications:
\begin{equation}
    \begin{aligned}
		\boldsymbol{z}_L = \sum_{i=1}^q \beta_{i,L}(\boldsymbol{u}) \cdot \tau_i (\boldsymbol{x}), \\
		\boldsymbol{z}_U = \sum_{i=1}^q \beta_{i,U}(\boldsymbol{u}) \cdot \tau_i (\boldsymbol{x}).
	\end{aligned}
\end{equation}
The results from the two multiplications are then concatenated,  $\boldsymbol{z}^\dagger = [z_{1,L}, z_{1,U}, \ldots, z_{q,L}, z_{q,U}]$, and passed through a dense layer to produce output $\mathcal{G}_L (\boldsymbol{u})(\boldsymbol{x})$ and $\mathcal{G}_U (\boldsymbol{u})(\boldsymbol{x})$.

\subsection{Bound propagation method}
\label{sec: bp_dip}

Rooted in the fields of neural network verification and certified adversarial defense, bound propagation methods aim to train deep neural networks that are provably robust to bounded input perturbations~\cite{gowal2019}. Prominent techniques in this category include interval bound propagation (IBP)~\cite{gowal2019}, DeepPoly~\cite{Singh2019}, and CROWN along with its various extensions~\cite{zhang2018, zhang2019, wang2021beta}. Although these methods differ in their underlying formulations, they share a common objective: to propagate interval-valued inputs through the network in order to obtain certified bounds on the output, thereby enhancing model robustness.

Inspired by this line of research, we adapt bound propagation techniques to address interval uncertainty in physical models, employing neural networks as surrogate models. For our experiments, we use the \texttt{Auto\_LiRPA} Python package~\cite{xu2020}, which provides a flexible interface for applying bound propagation to standard neural network architectures. Among the available methods, we focus on the standard implementations of IBP and CROWN. IBP is selected for its simplicity and memory efficiency, making it suitable for large-scale problems, while CROWN is chosen for its ability to produce tighter output bounds~\cite{zhang2019}. However, we observed that the \texttt{Auto\_LiRPA} implementation of CROWN requires substantially more memory than IBP.

\subsubsection{IBP}
\label{sec: ibp}
Stemming from neural network verification, interval bound propagation (IBP)~\cite{gowal2019} provide a simple bounding technique that propagates the perturbation in the input space to produce bounded outputs. Consider a standard feedforward neural network defined as a composition of layer-wise transformations:
\begin{equation}
    \hat{f}(\boldsymbol{x}) = z_M \circ z_{M-1} \circ \cdots \circ z_1(\boldsymbol{x}),
    \label{eq: nn}
\end{equation}
where each layer $ z_l, \quad l = 1,\ldots,M $ is given by:
\begin{equation}
    z_l = \sigma(\hat{z}_l), \quad \text{with} \quad \hat{z}_l = \boldsymbol{W}_l z_{l-1} + \boldsymbol{b}_l,
    \label{eq: act_fun}
\end{equation}
with $ \boldsymbol{W}_l \in \mathbb{R}^{d_l \times d_{l-1}} $ denoting the weight matrix, $ \boldsymbol{b}_l \in \mathbb{R}^{d_l} $ the bias vector, and $ \sigma(\cdot) $ an element-wise activation function. For clarity and without loss of generality, we assume the ReLU activation, $ \sigma(z) = \max(0, z) $. In this paper, we use the notation $ \hat{z}_l $ to represent the \emph{pre-activation} values of the $l$-th layer and $ z_l $ to represent the \emph{post-activation} values.

In IBP, we are interested in verifying that a NN satisfies specifications that for all inputs in some set $\mathcal{X}(x_0)$ around $x_0$, the network output satisfies a linear relationship
\begin{equation}
    c^\intercal z_l + d \leq 0 \quad \forall z_0 \in \mathcal{X}(x_0),
\end{equation}
where $c$ and $d$ are a vector and a scalar that may depend on the nominal input $x_0$ and label $y_{\text{true}}$. Specifically, in IBP we focus on the robustness to adversarial perturbation within $\ell_\infty$ norm-bounded ball around input $x_0$. We can verify the specification by searching counter-example that violates the specification constraint:
\begin{equation}
\begin{aligned}
    \max_{z_0 \in \mathcal{X}(x_0)} \quad & c^\intercal z_l + d \\
    \text{s.t.} \quad & z_l = \sigma(\boldsymbol{W}_l z_{l-1} + \boldsymbol{b}_l), \ l = 1,\ldots,L. 
\end{aligned}
\label{eq:spec_const}
\end{equation}
If the optimal value of the optimisation problem is less than 0, then the specification is satisfied.

The goal of IBP is to find the upper bound on the optimal value on problem~\ref{eq:spec_const}. The simplest approach is to bound the activation of each layer $z_l$ with an axis-aligned bounding box. Given $e_i$ is the standard $i^{\text{th}}$ basis vector and $\epsilon$ is the perturbation radius, for $\ell_\infty$ perturbation of size $\epsilon$, we have for each coordinate $z_{l,i}$ of $z_l$:
\begin{equation}
\begin{aligned}
    z_{L_{l,i}} (\epsilon) = \min_{z_{L_{l-1}} (\epsilon) \leq z_{l-1} \leq z_{U_{l-1}} (\epsilon)} e_i^\intercal \sigma(\boldsymbol{W}_l z_{l-1} + \boldsymbol{b}_l), \\
    z_{U_{l,i}} (\epsilon) = \max_{z_{L_{l-1}} (\epsilon) \leq z_{l-1} \leq z_{U_{l-1}} (\epsilon)} e_i^\intercal \sigma(\boldsymbol{W}_l z_{l-1} + \boldsymbol{b}_l),
\end{aligned}
\end{equation}
where subscript $L$ and $U$ denotes the lower and upper bound of $z$. In the input layer the pertubation is defined as $z_{L_0} = x_0 - \epsilon \boldsymbol{1}$ and $z_{U_0} = x_0 + \epsilon \boldsymbol{1}$. The computation in affine layers, such as fully-connected layers and convolutions can be done effectively with:
\begin{equation}
\begin{aligned}
    \mu_{l-1} = & \frac{z_{U_{l-1}} + z_{L_{l-1}}}{2} \\
    r_{l-1} = & \frac{z_{U_{l-1}} - z_{L_{l-1}}}{2} \\
    \mu_l = & \boldsymbol{W}_l \mu_{l-1} + \boldsymbol{b}_l \\
    r_l = & |\boldsymbol{W}_l| r_{l-1} \\
    z_{L_l} = & \sigma(\mu_l - r_l) \\
    z_{U_l} = & \sigma(\mu_l + r_l),
\end{aligned}
\end{equation}
where $|\cdot|$ is the element-wise absolute value operation, and propagating bounds through any element-wise monotonic function such as ReLU, sigmoid, and tanh is trivial. In contrast to the original paper~\cite{gowal2019}, the problem that are discussed in this paper are regression. Therefore, we define our loss function given in section~\ref{sec: BP_implement}.

\subsubsection{CROWN}
\label{sec: crown}
Neural networks are inherently nonlinear due to the presence of activation functions, such as the Rectified Linear Unit (ReLU), applied at each layer. This nonlinearity presents a major challenge in computing certified output bounds for neural networks under input perturbations. To address this, the CROWN method introduces a linear relaxation framework that propagates bounds through the network in a layer-wise fashion~\cite{zhang2018}, enabling tractable and certified bound estimation for adversarial robustness and uncertainty quantification.

Let us recall the same neural network notation in equations~\ref{eq: nn} and~\ref{eq: act_fun}. To certify the output under input perturbations $ \boldsymbol{x} \in \mathcal{X} $, where $ \mathcal{X} = \{\boldsymbol{x} \in \mathbb{R}^{d_0} \mid \boldsymbol{x}_{0,L} \leq \boldsymbol{x} \leq \boldsymbol{x}_{0,U} \} $, CROWN propagates linear upper and lower bounds through the network. At each layer, given a lower and upper pre-activation bound $ \hat{z}_{l,L} \leq \hat{z}_l \leq \hat{z}_{l,U} $, the ReLU activation is relaxed using two linear constraints: a linear upper bound 
\begin{equation}
z_l \leq \frac{\hat{z}_{l,U}}{\hat{z}_{l,U} - \hat{z}_{l,L}} \left(\hat{z}_l - \hat{z}_{l,L} \right),    
\end{equation}
and a linear lower bound
\begin{equation}
    z_l \geq \boldsymbol{\alpha}_l \hat{z}_l, \quad (0 \leq \boldsymbol{\alpha}_l \leq 1),
\end{equation}

where $ \boldsymbol{\alpha}_l $ define the slope of the linear lower bound on the ReLU function over the interval $ [\hat{z}_{l,L}, \hat{z}_{l,U}] $.

These linear bounds allow one to propagate the convex outer approximations of the activation functions through the entire network. Given the piecewise linearity, the output of the network can now be bounded by solving a pair of linear programs (LPs) corresponding to the lower and upper bounds:

\begin{equation}
    f^L(\boldsymbol{x}) \leq \hat{f}(\boldsymbol{x}) \leq f^U(\boldsymbol{x}), \quad \forall \boldsymbol{x} \in \mathcal{X},
\end{equation}

where $ f_L(\cdot) $ and $ f_U(\cdot) $ are affine functions of the input $ \boldsymbol{x} $, recursively defined by the linear relaxation parameters of the activations at each layer. Moreover, the relaxation can also be applied to other activation functions, making CROWN adaptable to many neural network architectures.

Adapting the CROWN algorithm for interval bound propagation within the DeepONet architecture is relatively straightforward, as it does not require modifications to the original network design. Leveraging the \texttt{Auto\_LiRPA} library~\cite{xu2020}, bound propagation in DeepONet can be implemented directly by using the standard DeepONet architecture (as illustrated in Figure~\ref{fig: deeponet}) in combination with the bound propagation wrapper provided by the library.

Suppose we are given a dataset consisting of $n$ pairs of interval-valued input-output functions arising from a PDE system:
\[
\left\{ \left[ \boldsymbol{u}_{1,L}, \boldsymbol{u}_{1,U} \right], \ldots, \left[ \boldsymbol{u}_{n,L}, \boldsymbol{u}_{n,U} \right] \right\}, \quad
\left\{ \left[ \boldsymbol{g}_{1,L}, \boldsymbol{g}_{1,U} \right], \ldots, \left[ \boldsymbol{g}_{n,L}, \boldsymbol{g}_{n,U} \right] \right\},
\]
where $\boldsymbol{u}_{i,L}$ and $\boldsymbol{u}_{i,U}$ denote the lower and upper bounds of the $i$-th input function, and likewise for the output functions.

To interface with \texttt{Auto\_LiRPA}, each interval-valued input is transformed into its center and perturbation form:
\begin{equation}
    \begin{aligned}
        \boldsymbol{u}_c &= \frac{\boldsymbol{u}_U + \boldsymbol{u}_L}{2}, \\
        \boldsymbol{u}_p &= \frac{\boldsymbol{u}_U - \boldsymbol{u}_L}{2},
    \end{aligned}
\end{equation}
where $ \boldsymbol{u}_c $ represents the nominal input (center) and $ \boldsymbol{u}_p $ the element-wise perturbation radius.

\subsubsection{Bound propagation DeepONet implementation}
\label{sec: BP_implement}

The standard DeepONet model is extended to handle interval-valued data through a bound propagation framework. During training, the wrapped model operates within a conventional optimisation loop, receiving as input the pair of branch and trunk encodings, denoted by $[ \boldsymbol{u}_c, \boldsymbol{u}_p ]$. The model is trained to predict interval bounds $[ \hat{\boldsymbol{g}}_L, \hat{\boldsymbol{g}}_U ]$ that approximate the ground-truth output intervals $[ \boldsymbol{g}_L, \boldsymbol{g}_U ]$. A natural baseline for this setting is the \textit{bound-based loss}, defined as the sum of mean squared errors (MSE) between the predicted and true lower and upper bounds, together with a non-crossing regularisation term:
\begin{equation}
    \mathcal{L}_{\text{bound}} = 
    \text{MSE}(\boldsymbol{g}_L, \hat{\boldsymbol{g}}_L)
    + \text{MSE}(\boldsymbol{g}_U, \hat{\boldsymbol{g}}_U)
    + \lambda \left( \max(0, \hat{\boldsymbol{g}}_L - \hat{\boldsymbol{g}}_U)^2 \right),
    \label{eq:intv_loss}
\end{equation}
where $\lambda$ is a regularisation coefficient that enforces the non-crossing constraint between the predicted lower and upper bounds. This formulation explicitly penalises errors in both interval endpoints and serves as a direct extension of classical regression objectives to interval-valued targets.

However, in certain scenarios, this bound-based approach can be overly restrictive, particularly when the true interval labels are noisy or when the model has limited expressive capacity. Explicitly fitting both upper and lower bounds may cause the model to overfit uncertain boundaries instead of capturing the underlying functional relationship. To address this limitation, we introduce an alternative objective, referred to as the \textit{midpoint-based loss}. This formulation focuses on the midpoint of the interval rather than its individual bounds, allowing the model to learn the central trend of the data while still maintaining the non-crossing constraint:
\begin{equation}
    \boldsymbol{g}_M = \frac{1}{2}(\boldsymbol{g}_L + \boldsymbol{g}_U), \quad
    \hat{\boldsymbol{g}}_M = \frac{1}{2}(\hat{\boldsymbol{g}}_L + \hat{\boldsymbol{g}}_U),
\end{equation}
\begin{equation}
    \mathcal{L}_{\text{mid}} =
    \text{MSE}(\boldsymbol{g}_M, \hat{\boldsymbol{g}}_M) + \lambda \left( \max(0, \hat{\boldsymbol{g}}_U - \hat{\boldsymbol{g}}_L)^2 \right).
    \label{eq:mid_loss}
\end{equation}

The midpoint-based formulation relaxes the constraints imposed by direct bound matching, which can improve training stability and reduce sensitivity to interval width or asymmetry. This flexibility is advantageous in practical engineering problems where the uncertainty bounds are estimated or partially known rather than measured precisely.

In this study, both loss functions in Eqs.~\eqref{eq:intv_loss} and~\eqref{eq:mid_loss} are systematically compared to assess their respective trade-offs between conservativeness, stability, and accuracy. The bound-based loss emphasises tight interval fitting, while the midpoint-based loss emphasises capturing the central tendency of the target distribution. 

Furthermore, for applications where conservative predictions are desired, such as in aerospace design where underpredicting aerodynamic loads or thermal stresses can compromise safety margins and result in catastrophic system damage, an asymmetric loss function can be employed to replace MSE in Eq.~\ref{eq:intv_loss}. One suitable choice is the linear-exponential (Linex) loss~\cite{Varian1975,Fu2023}, defined as
\begin{equation}
    \text{Linex}(y, \hat{y}) = \exp(a(y - \hat{y})) - a(y - \hat{y}) - 1,
    \label{eq:linex_loss}
\end{equation}
where $a$ is an asymmetry parameter. Larger values of $a$ increase the penalty for underestimation relative to overestimation, thereby promoting conservative predictions. Replacing the MSE term in either Eq.~\eqref{eq:intv_loss} or Eq.~\eqref{eq:mid_loss} with the Linex loss allows the model to learn tighter yet safety-oriented intervals. This loss formulation is applicable not only to the bound propagation method but also to naive direct interval propagation (DIP) schemes and interval neural network architectures.

\subsection{Interval neural-network method}
\label{sec: inn_dip}

\subsubsection{INN standard implementation}
In contrast to the naive approach introduced in section \ref{sec: naive_dip}, the interval neural network (INN) term here refers to a neural network model which network weights and biases are interval-valued~\cite{Ishibuchi1993,Oala2021,Betancourt2022,Tretiak2023,Wang2025}. Therefore, the model is able to handle interval-valued input and produce interval-valued output. A forward propagation in the $l$-th layer in the INN is expressed as:
\begin{equation}
    [\boldsymbol{z}_L, \boldsymbol{z}_U]_l = \sigma\left( [\boldsymbol{W}_L,\boldsymbol{W}_U]_l \odot [\boldsymbol{z}_L, \boldsymbol{z}_U]_{l-1} \oplus [\boldsymbol{b}_L, \boldsymbol{b}_U]_l \right),
    \label{eq: inn_forward}
\end{equation}
where $\odot$, $\oplus$, and $\ominus$ are interval multiplication, addition and subtraction, defined as:
\begin{equation}
    \begin{aligned}
      \relax [a_L, a_U] \oplus [a_L, a_U] = &[a_L + a_L, a_U + a_U]\\
      [a_L, a_U] \ominus [a_L, a_U] = &[a_L - a_U, a_U - a_L]\\
      [a_L, a_U] \odot [a_L, a_U] = & [\min(a_L \cdot a_L, a_L \cdot a_U, a_U \cdot a_L, a_U \cdot a_U) , \\
      & \max(a_L \cdot a_L, a_L \cdot a_U, a_U \cdot a_L, a_U \cdot a_U)].\\
    \end{aligned}
    \label{eq: interval_algebra}
\end{equation}

By using some reformulation tricks, the smoothness of equation \ref{eq: inn_forward} can be guaranteed. By analysing the interval multiplication under various conditions (including negative value) of $\boldsymbol{z}_L, \boldsymbol{z}_U, \boldsymbol{W}_L,$ and $\boldsymbol{W}_U$, the multiplication in equation \ref{eq: inn_forward}, denoted as $[c_L, c_U] := [\boldsymbol{W}_L,\boldsymbol{W}_U] \odot [\boldsymbol{z}_L, \boldsymbol{z}_U]$ can be rewritten as:
\begin{equation}
    \begin{aligned}
        c_L = &\min\{\boldsymbol{W}_U,0\} \min\{\boldsymbol{z}_U,0\} + \max\{\boldsymbol{W}_U,0\} \max\{\boldsymbol{z}_U,0\} \\
        & + \min\{ \max\{\boldsymbol{W}_U,0\} \min\{\boldsymbol{z}_L,0\} - \min\{\boldsymbol{W}_L,0\} \max\{\boldsymbol{z}_U,0\},0\}\\
        & + \min\{\boldsymbol{W}_L,0\} \max\{\boldsymbol{z}_U,0\}, \\
        c_U = &\min\{\boldsymbol{W}_U,0\} \max\{\boldsymbol{z}_L,0\} + \max\{\boldsymbol{W}_L,0\} \min\{\boldsymbol{z}_U,0\} \\
        & + \max\{ \min\{\boldsymbol{W}_L,0\} \min\{\boldsymbol{z}_L,0\} - \max\{\boldsymbol{W}_U,0\} \max\{\boldsymbol{z}_U,0\},0\}\\
        & + \max\{\boldsymbol{W}_U,0\} \max\{\boldsymbol{z}_U,0\}.
    \end{aligned}
    \label{eq: smooth_inn}
\end{equation}
Although the operations in equation \ref{eq: smooth_inn} are not strictly differentiable at zeros, it can be observed that the min and max operations are continuous. Thus, the smoothness of the forward propagation ensures that the parameter updates via backpropagation through automatic differentiation as a standard neural network is attainable~\cite{Oala2021}. 

In case of using non-negative activation function such as the ReLU activation, the forward propagation in equation \ref{eq: inn_forward} is simplified as:
\begin{equation}
    \begin{aligned}
        \boldsymbol{z}_{l,L} &\!=\!\sigma^{l}\bigg(\! \operatorname*{min}\{\boldsymbol{W}_{l,L}, \boldsymbol{0}\} \boldsymbol{a}_{l-1,U} + \operatorname*{max}\{\boldsymbol{W}_{l,L}, \boldsymbol{0}\}\boldsymbol{a}_{l-1,L} + \boldsymbol{b}_{l,L} \bigg) \\  
        \boldsymbol{z}_{l,U} &\!=\!\sigma^{l}\bigg(\! \operatorname*{max}\{\boldsymbol{W}_{l,U}, \boldsymbol{0}\}\boldsymbol{a}_{l-1,U} + \operatorname*{min}\{\boldsymbol{W}_{l,U}, \boldsymbol{0}\}\boldsymbol{a}_{l-1,L}+ \boldsymbol{b}_{l,U} \bigg)
    \end{aligned}
\end{equation}

To integrate an Interval Neural Network (INN) into the DeepONet architecture, we adopt a strategy analogous to the one used for naive Direct Interval Propagation (naive DIP), as described in Section~\ref{sec: naive_dip}. Assuming that only the branch input is interval-valued, we modify the standard DeepONet (illustrated in Figure~\ref{fig: deeponet}) by replacing the fully connected neural network (FNN) in the branch net with an INN. This substitution enables the network to handle interval-valued inputs in a mathematically principled manner.

\subsubsection{INN DeepONet implementation}
In contrast to the DeepONet with naive DIP, we introduce an \textit{interval multiplication layer} to combine the outputs of the branch and trunk nets. This layer performs element-wise interval multiplication, as defined in Equation~\ref{eq: interval_algebra}, and is designed to propagate interval uncertainty through the network. While we assume interval inputs only in the branch net, this formulation also opens the possibility for the trunk net to be interval-valued, offering greater modelling flexibility.

\begin{figure}[ht]
    \begin{center}
    \includegraphics[width=.65\textwidth]{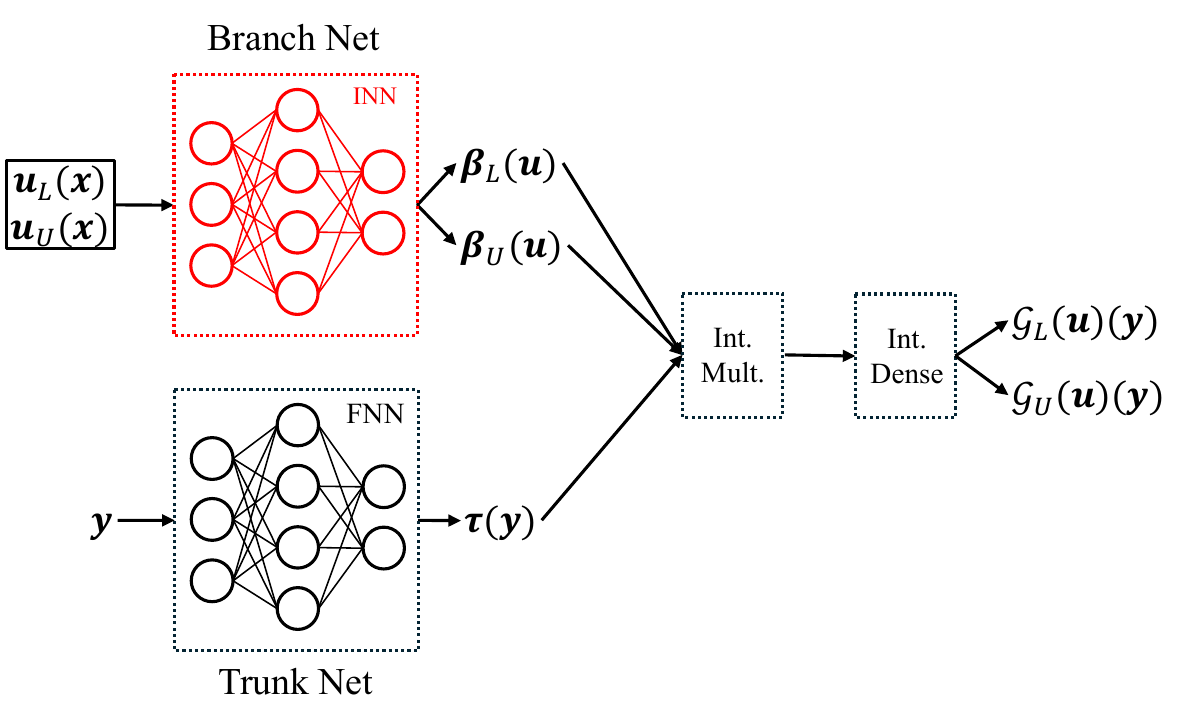}
    \end{center}
\vspace{-0.7cm}
\caption{Interval neural network (INN) with DeepONet architecture.}
\label{fig: inn_deeponet}
\end{figure}

The output of the interval multiplication layer is subsequently passed through an \textit{interval dense layer}, which performs standard interval forward propagation (as defined in Equation~\ref{eq: inn_forward}) using a linear activation function.  The model is trained using a loss function defined as the sum of mean squared errors (MSE) between the predicted and ground-truth lower and upper output bounds, as given in Equation~\ref{eq:intv_loss}. This loss encourages the network to produce interval predictions that tightly enclose the true target function. The overall architecture of DeepONet with INN is depicted in Figure~\ref{fig: inn_deeponet}.

\subsection{Connection to interval estimation methods}
\label{sec: compare_intv}
Interval prediction/estimation methods aim to quantify predictive uncertainty by providing interval-valued outputs that capture a range of plausible outcomes. Following~\cite{Abdi2023, Dai2024}, we distinguish between uncertainty prediction and uncertainty propagation. Common interval prediction approaches include quantile regression~\cite{Koenker1978, Steinwart2011, Pouplin2025} and conformal prediction~\cite{Gammerman2005CP,Papadopoulos2002CP}. These methods typically estimate prediction intervals based on statistical properties of the data, rather than explicitly propagating input uncertainty through a deterministic model. Consequently, they primarily focus on achieving desirable statistical properties, such as coverage probability and interval efficiency, instead of modelling the propagation of interval-valued inputs. This represents a fundamental difference in objective compared to the interval propagation methods considered in this paper.
Quantile regression, for example, aims to achieve probabilistic coverage of the target variable by directly estimating specified quantiles of the conditional output distribution. To this end, it is trained using the quantile (pinball) loss function:
\begin{equation}
\mathcal{L}_{\text{quantile}} = \sum_{i=1}^{n} \left[ \tau \cdot \max(0, y_i - \hat{y}_i) + (1 - \tau) \cdot \max(0, \hat{y}_i - y_i) \right],
\end{equation}
where $\tau \in (0,1)$ denotes the target quantile level (e.g., $\tau = 0.05$ corresponds to the 5th percentile).
Conformal prediction is often applied as a model-agnostic post-processing procedure that calibrates prediction intervals to achieve a user-specified coverage probability under mild distributional assumptions. Unlike quantile regression, conformal methods do not directly model the conditional distribution, but instead adjust interval widths using calibration data to provide finite-sample coverage guarantees.

In contrast, interval propagation methods focus on the deterministic transformation of input intervals through a system or model to obtain output intervals that reflect propagated uncertainty. The associated loss functions, such as those defined in Equations~\ref{eq:intv_loss}, are formulated to minimise the discrepancy between predicted and reference interval bound rather than achieving a prescribed statistical coverage. In certain safety-critical applications, conservative predictions may be preferred. For example, in aerospace design, underpredicting aerodynamic loads or thermal stresses can compromise safety margins and lead to severe system failure. In such cases, asymmetric loss functions, such as the Linex loss defined in Equation~\ref{eq:linex_loss}, can be employed to penalise underestimation more strongly than overestimation, thereby encouraging conservative interval predictions. 

Despite their conceptual differences, interval estimation and interval propagation methods can be complementary. Interval propagation may provide deterministic bounds that can subsequently be refined using statistical calibration techniques to improve coverage and efficiency. Conversely, interval estimation methods can offer insights into the statistical characteristics of the data that may inform the design of interval propagation models. Although both approaches produce interval-valued predictions, they address distinct objectives and rely on different methodological frameworks. The integration of interval estimation and interval propagation techniques is reserved for future work.

Nevertheless, evaluating coverage metrics remains informative when assessing interval propagation methods, particularly when the reference interval labels are noisy or indirectly estimated. Therefore, in this paper we also provide the interval coverage metrics such as predicted interval normalized average width (PINAW), predicted interval coverage probability (PICP), and the coverage width-based criterion (CWC) defined as:
\begin{equation}
    \text{PINAW} = \frac{1}{N} \sum_{i=1}^N \frac{(\hat{y}_U - \hat{y}_L)}{(y_U - y_L)},
\end{equation}
\begin{equation}
    \text{PICP} = \frac{1}{N} \sum_{i=1}^N \boldsymbol{1}\left( \hat{y}_L \leq y_i \leq \hat{y}_U \right),
\end{equation}
\begin{equation}
    \text{CWC} = \text{PINAW}\left( 1 + e^{\gamma \max(0,(1-\delta)-\text{PICP})} \right),
\end{equation}
where $\boldsymbol{1}(\cdot)$ is a count function that returns one if event A occurs and zero otherwise, $\delta$ in CWC represent the desired error-rate, and $\gamma$ is a user-defined parameters that balances the trade-off between two objectives. In our case we use $\delta = 0$ since we would like to the predicted interval covers all the actual interval, and $\gamma = 5$ to produce reasonable value between the objectives. However, since we want to evaluate the prediction with respect to the actual interval bound, we modify the PICP to use the continuous overlap-based coverage:
\begin{equation}
    \text{PICP} = \frac{1}{N} \sum_{i=1}^N \frac{\max(0,\min(\hat{y}_U,y_U ) - \max(\hat{y}_L, y_L))}{(y_U - y_L)}.
\end{equation}

\section{Experiments}
\label{sec: experiments}

To evaluate and compare the performance of the three proposed methods, we design two types of experimental setups: one using ideal interval-valued data and the other using pointwise (precise) data. The first experiment assumes access to ground-truth interval-valued inputs and outputs, which represent the ideal scenario for interval learning. However, such datasets are rarely available in practice, as most engineering computational simulations typically produce pointwise data. That is, for a given input configuration, the simulation or experiment yields a single deterministic output.

To address this limitation, our second experiment utilises an augmented pointwise dataset. Specifically, we construct synthetic interval-valued data by augmenting the original pointwise samples with controlled perturbations. This allows us to assess the robustness and generalisability of the models under uncertainty, even in the absence of true interval labels. The details of the augmentation strategy used to generate the interval dataset from pointwise data are provided in a dedicated subsection later in this paper. 

In this numerical study, we compare seven methods, including the optimisation-based interval propagation baseline (Opt-Prop). The evaluated approaches comprise the naive direct interval propagation and interval neural networks (INN), described in Sections~\ref{sec: naive_dip} and~\ref{sec: inn_dip}, respectively. We further include bound-propagation techniques, namely IBP and CROWN, as introduced in Sections~\ref{sec: ibp} and~\ref{sec: crown}. In addition, midpoint variants of IBP and CROWN are considered using the loss formulation defined in equation~\ref{eq:mid_loss}. Direct comparison with the Opt-Prop baseline is not always feasible, particularly for PDE-based test cases, due to its prohibitive computational cost. Nevertheless, its computational complexity can be approximately inferred by extrapolating the scaling behaviour observed in simpler benchmark problems. 

Training hyperparameters, including the regularisation coefficient $\lambda$, learning rate, and number of training epochs, are tuned separately for each model and test case to achieve a balance between predictive performance and training efficiency. To ensure a consistent and fair comparison of computational cost, the neural network architecture is fixed to 16 nodes with 3 hidden layers in the one-dimensional regression problem,  64 nodes with 4 hidden layers for both trunk- and branch-net in the one-dimensional PDE problem, 128 nodes with 2 hidden layers for trunk-net and 128 nodes with 1 hidden layer for branch-net in the two-dimensional PDE problem. Training time is reported by measuring the elapsed time up to a predefined epoch count. Specifically, the timer is evaluated at 2000 epochs for the one-dimensional regression problem, 1000 epochs for the one-dimensional PDE problem, and 150 epochs for the two-dimensional PDE problem. 

\subsection{Experiments with ideal interval data}
\label{sec: exp_ideal}

In this experimental setting, we assume full access to ground-truth interval-valued inputs and outputs, making it an ideal case for evaluating the methods under controlled conditions. As a preliminary validation, we first apply all three methods to a simple one-dimensional regression task, serving as a basic verification to ensure that each model is capable of propagating interval uncertainty correctly in a regression problem. Following this, we evaluate the methods on a basic one-dimensional PDE problem using the DeepONet framework. Finally, we extend the assessment to a more complex two-dimensional PDE problem to examine the scalability and effectiveness of the methods in higher-dimensional settings.

\subsubsection{1-dimensional regression}
\label{sec: 1dreg_ideal}

\begin{figure}[ht]
    \begin{center}
    \includegraphics[width=.55\textwidth]{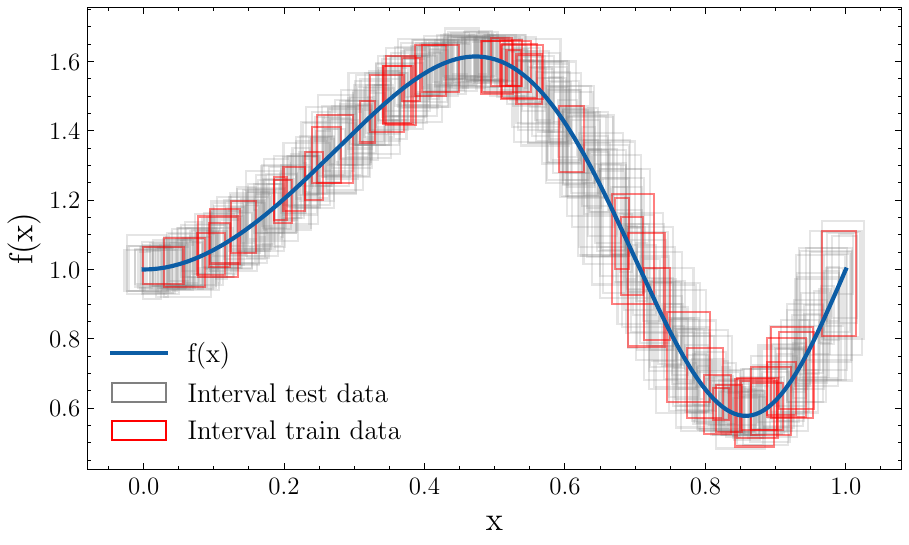}
    \end{center}
\vspace{-0.7cm}
\caption{Ideal interval dataset for a one-dimensional regression problem. Single rectangle represent a single pair of interval-valued $x$ and $y$.}
\label{fig: ideal_1dreg}
\end{figure}

Consider a simple one-dimensional regression problem:
\begin{equation}
    y = \sin(2x)e^{-x} + 1 + \varepsilon, \  x\in[0,\pi],
    \label{eq: 1dreg_problem}
\end{equation}
where $\varepsilon$ is a normally distributed aleatoric noise with mean $\mu=0$ and standard deviation $\sigma=0.025$. We create interval inputs with varying interval widths across the sample space. Since the actual function is cheap to evaluate, we propagate the inputs by evaluating on the actual function. Obtaining an interval dataset, illustrated in Figure~\ref{fig: ideal_1dreg}.

We evaluate each method across multiple training set sizes to examine its robustness under varying data availability. Since the problem is relatively simple, the neural network architecture is not expected to play a major role. Nevertheless, we include a brief hyperparameter study in Table~\ref{tab: ablation_1Dreg} for the two best-performing methods for completeness. For each configuration, the experiments are repeated ten times with different random seeds to account for stochastic variations in model initialisation and optimisation. The direct interval method is then compared to the baseline, denoted as Opt-Prop, which applies the optimisation-based interval propagation method using a surrogate model for comparison. The Opt-Prop method constructs a standard surrogate model on pointwise data, and then for each $\boldsymbol{x}_i$ interval we solve the optimisation problem defined in equation~\ref{eq: interval_prop}. In this setting, we use \texttt{COBYLA} optimiser provided in \texttt{scipy} library. In the ideal interval data setting, the surrogate model in Opt-Prop method is trained on the center points of the interval data.

In interval regression tasks, prediction errors are commonly decomposed into lower- and upper-bound components to assess how well models capture the uncertainty range. As shown in Table~\ref{tab: ideal_1dreg}, with sufficient training data (25 samples), all methods except mid-IBP outperform the Opt-Prop baseline in terms of RMSE. Coverage metrics results in Table~\ref{tab: ideal_1dreg_metrics} indicate that Opt-Prop consistently under-covers the true bounds, leading to low PICP values. This behaviour is expected, as Opt-Prop relies on a surrogate that predicts only the mean response and neglects aleatoric noise. In contrast, direct interval propagation methods more effectively capture the true bounds by directly modelling interval-valued data. To address the fairness of this comparison, we further consider a more realistic setting in Section~\ref{sec: 1dreg_aug}, where only pointwise training data are available.

Computational costs, reported as inference times in Table~\ref{tab: ideal_1dreg_time}, show that even the slowest direct interval propagation method (INN) is approximately three orders of magnitude faster than Opt-Prop. This substantial speedup underscores the scalability advantages of direct interval propagation, particularly for large-scale or time-sensitive applications.

\begin{table}[th]
    \centering
    \caption{Ideal 1-dimensional regression RMSE comparison across methods and training set sizes.}
    {\small
    \setlength{\tabcolsep}{4pt}
    \begin{tabular}{c c | c c c c c c c}
    \toprule
    \textbf{$n_\text{train}$} & \textbf{Metric} & \textbf{Naive} & \textbf{IBP} & \textbf{CROWN} & \textbf{INN}  & \textbf{Mid-CROWN} & \textbf{Mid-IBP} & \textbf{Opt-Prop}\\
    \midrule
    \multirow{2}{*}{10}  & $\text{RMSE}_L$ & 0.076 $\pm$ 0.047 & 0.079 $\pm$ 0.031 & 0.107 $\pm$ 0.067 & 0.085 $\pm$ 0.051 & 0.049 $\pm$ 0.017 & 0.323 $\pm$ 0.086  & 0.053 $\pm$ 0.010 \\
                        & $\text{RMSE}_U$ & 0.136 $\pm$ 0.102 & 0.087 $\pm$ 0.057 & 0.109 $\pm$ 0.057 & 0.125 $\pm$ 0.177 & 0.059 $\pm$ 0.022 & 0.332 $\pm$ 0.113 & 0.063 $\pm$ 0.027 \\
    \hline 
    \multirow{2}{*}{25}  & $\text{RMSE}_L$ & 0.025 $\pm$ 0.008 & 0.035 $\pm$ 0.006 & 0.066 $\pm$ 0.030 & 0.035 $\pm$ 0.030 & 0.040 $\pm$ 0.002 & 0.291 $\pm$ 0.084 & 0.054 $\pm$ 0.006 \\
                        & $\text{RMSE}_U$ & 0.038 $\pm$ 0.027 & 0.033 $\pm$ 0.006 & 0.059 $\pm$ 0.013 & 0.037 $\pm$ 0.035 & 0.041 $\pm$ 0.002 & 0.295 $\pm$ 0.083 & 0.045 $\pm$ 0.004\\
    \hline 
    \multirow{2}{*}{50} & $\text{RMSE}_L$ & 0.023 $\pm$ 0.005 & 0.029 $\pm$ 0.004 & 0.033 $\pm$ 0.005 & 0.019 $\pm$ 0.004 & 0.040 $\pm$ 0.004 & 0.297 $\pm$ 0.044 & 0.043 $\pm$ 0.007\\
                        & $\text{RMSE}_U$ & 0.026 $\pm$ 0.006 & 0.027 $\pm$ 0.002 & 0.049 $\pm$ 0.009 & 0.019 $\pm$ 0.003 & 0.040 $\pm$ 0.004 & 0.299 $\pm$ 0.045 & 0.053 $\pm$ 0.006\\
    \hline
    \multirow{2}{*}{100} & $\text{RMSE}_L$ & 0.020 $\pm$ 0.004 & 0.030 $\pm$ 0.004 & 0.034 $\pm$ 0.005 & 0.018 $\pm$ 0.003 & 0.041 $\pm$ 0.005 & 0.190 $\pm$ 0.039 & 0.055 $\pm$ 0.006 \\
                        & $\text{RMSE}_U$ & 0.021 $\pm$ 0.005 & 0.031 $\pm$ 0.003 & 0.041 $\pm$ 0.004 & 0.019 $\pm$ 0.005 & 0.040 $\pm$ 0.004 & 0.191 $\pm$ 0.036 & 0.046 $\pm$ 0.005 \\
    \bottomrule
    \end{tabular}
    }
    \label{tab: ideal_1dreg}
\end{table}

\begin{table}[th]
\centering
\caption{Ideal 1-dimensional regression performance comparison across methods and training set sizes on three different metrics: predicted interval normalized average width (PINAW), predictied interval coverage probability (PICP), and the coverage width-based criterion (CWC).}
{\small
\setlength{\tabcolsep}{4pt}
\begin{tabular}{c c | c c c c c c c}
\toprule
\textbf{$n_\text{train}$} & \textbf{Metric} & \textbf{Naive} & \textbf{IBP} & \textbf{CROWN} & \textbf{INN} & \textbf{Mid-CROWN} & \textbf{Mid-IBP} & \textbf{Opt-Prop}\\
\midrule
\multirow{3}{*}{10}  & PINAW & $0.999 \pm 0.241$ & $1.069 \pm 0.198$ & $1.064 \pm 0.196$ & $1.186 \pm 0.756$ & $0.620 \pm 0.057$ & $3.316 \pm 0.215$ & $0.477 \pm 0.013$ \\
                     & PICP & $0.742 \pm 0.117$ & $0.758 \pm 0.060$ & $0.657 \pm 0.091$ & $0.758 \pm 0.144$ & $0.558 \pm 0.040$ & $0.982 \pm 0.012$ & $0.431 \pm 0.057$ \\
                     & CWC & $5.287 \pm 3.442$ & $4.774 \pm 1.349$ & $7.571 \pm 3.560$ & $5.241 \pm 2.532$ & $6.395 \pm 1.366$ & $6.942 \pm 0.395$ & $9.066 \pm 3.031$ \\
\hline 
\multirow{3}{*}{25} & PINAW  & $1.060 \pm 0.082$ & $1.000 \pm 0.043$ & $0.932 \pm 0.142$ & $1.074 \pm 0.110$ & $0.567 \pm 0.021$ & $2.843 \pm 0.949$ & $0.475 \pm 0.011$\\
                     & PICP  & $0.889 \pm 0.045$ & $0.842 \pm 0.023$ & $0.706 \pm 0.032$ & $0.886 \pm 0.084$ & $0.553 \pm 0.017$ & $0.991 \pm 0.006$ & $0.465 \pm 0.008$ \\
                     & CWC & $2.981 \pm 0.689$ & $3.214 \pm 0.201$ & $4.946 \pm 0.468$ & $3.280 \pm 1.768$ & $5.862 \pm 0.254$ & $5.792 \pm 1.848$ & $7.385 \pm 0.158$ \\
\hline 
\multirow{3}{*}{50} & PINAW  & $1.045 \pm 0.034$ & $0.943 \pm 0.044$ & $0.819 \pm 0.070$ & $1.046 \pm 0.031$ & $0.569 \pm 0.010$ & $3.669 \pm 0.307$ & $0.468 \pm 0.007$ \\
                     & PICP  & $0.909 \pm 0.029$ & $0.834 \pm 0.017$ & $0.707 \pm 0.025$ & $0.932 \pm 0.026$ & $0.558 \pm 0.007$ & $0.997 \pm 0.002$ & $0.463 \pm 0.007$  \\
                     & CWC & $2.704 \pm 0.208$ & $3.109 \pm 0.054$ & $4.362 \pm 0.272$ & $2.523 \pm 0.164$ & $5.754 \pm 0.095$ & $7.386 \pm 0.629$ & $7.320 \pm 0.173$\\
\hline
\multirow{3}{*}{100} & PINAW  & $1.006 \pm 0.102$ & $0.930 \pm 0.021$ & $0.729 \pm 0.048$ & $1.036 \pm 0.050$ & $0.562 \pm 0.014$ & $2.872 \pm 0.433$ & $0.454 \pm 0.008$\\
                     & PICP  & $0.905 \pm 0.055$ & $0.814 \pm 0.015$ & $0.670 \pm 0.031$ & $0.928 \pm 0.036$ & $0.554 \pm 0.012$ & $0.998 \pm 0.002$ & $0.450 \pm 0.011$ \\
                     & CWC & $2.639 \pm 0.228$ & $3.291 \pm 0.153$ & $4.529 \pm 0.316$ & $2.532 \pm 0.173$ & $5.783 \pm 0.183$ & $5.771 \pm 0.845$ & $7.559 \pm 0.302$ \\
\bottomrule
\end{tabular}
}
\label{tab: ideal_1dreg_metrics}
\end{table}

\begin{table}[!ht]
    \centering
    \caption{Ideal 1-dimensional regression time comparison in seconds (s). The IBP and CROWN-based methods are implemented using \texttt{PyTorch} and \texttt{Auto\_LiRPA} library, meanwhile the Naive, INN, and Opt-Prop methods are implemented using \texttt{TensorFlow}. Therefore the training and inference time between the two groups are not directly comparable. All experiments were conducted on MacBook Pro with Apple M1 Pro chip.}
    {\footnotesize
    \setlength{\tabcolsep}{4pt}
    \begin{tabular}{c c | c c c c c c c}
    \toprule
    \textbf{$n_\text{train}$} & \textbf{Process} & \textbf{Naive} & \textbf{IBP} & \textbf{CROWN} & \textbf{INN} & \textbf{Mid-CROWN} & \textbf{Mid-IBP} & \textbf{Opt-Prop}
    \\
    \midrule
    \multirow{2}{*}{10}  & Training & 16.504 $\pm$ 0.594 & 3.433 $\pm$ 0.105 & 24.448 $\pm$ 0.897 & 67.159 $\pm$ 1.609 & 23.467 $\pm$ 0.685 & 3.190 $\pm$ 0.122 & 12.353 $\pm$ 0.623 \\
                        & Inference & 0.078 $\pm$ 0.025 & 0.001 $\pm$ 0.001 & 0.023 $\pm$ 0.001& 0.213 $\pm$ 0.146 & 0.023 $\pm$ 0.001 & 0.001 $\pm$ 0.001 & 217.283 $\pm$ 4.548 \\
    \hline 
    \multirow{2}{*}{25}  & Training & 16.134 $\pm$ 1.021 & 3.394 $\pm$ 0.127 & 25.425 $\pm$ 0.661 & 67.958 $\pm$ 0.213 & 26.134 $\pm$ 1.349 & 3.316 $\pm$ 0.129 & 12.832 $\pm$ 0.947 \\
                        & Inference & 0.075 $\pm$ 0.028 & 0.001 $\pm$ 0.001 & 0.024 $\pm$ 0.001 & 0.202 $\pm$ 0.156 & 0.023 $\pm$ 0.002 & 0.001 $\pm$ 0.001 & 218.125 $\pm$ 3.775 \\
    \hline 
    \multirow{2}{*}{50} & Training & 30.813 $\pm$ 2.057 & 6.222 $\pm$ 0.222 & 49.373 $\pm$ 2.017 & 126.532 $\pm$ 7.255 & 51.897 $\pm$ 1.070 & 6.216 $\pm$ 0.032 & 22.484 $\pm$ 1.201 \\
                        & Inference & 0.088 $\pm$ 0.066 & 0.001 $\pm$ 0.001 & 0.022 $\pm$ 0.001 & 0.283 $\pm$ 0.278 & 0.023 $\pm$ 0.001 & 0.001 $\pm$ 0.001 & 213.077 $\pm$ 7.404 \\
    \hline
    \multirow{2}{*}{100} & Training & 57.159 $\pm$ 2.719 & 11.849 $\pm$ 0.374 & 100.206 $\pm$ 2.447 & 244.754 $\pm$ 5.408 & 107.701 $\pm$ 0.465 & 11.764 $\pm$ 0.101 & 42.209 $\pm$ 2.166 \\
                        & Inference & 0.078 $\pm$ 0.039 & 0.001 $\pm$ 0.001 & 0.022 $\pm$ 0.002 & 0.190 $\pm$ 0.149 & 0.022 $\pm$ 0.001 & 0.001 $\pm$ 0.001 & 217.300 $\pm$ 8.076 \\
    \bottomrule
    \end{tabular}
    }
    \label{tab: ideal_1dreg_time}
\end{table}

\subsubsection{1-dimensional PDE}
\label{sec: 1dpde_ideal}
We consider the second-order steady-state stochastic Poisson equation in one dimension as our first test case. The stochastic differential equation (SDE) is formulated as:
\begin{equation}
    \begin{aligned}
        && \frac{d^2 g(x)}{dx^2} = 20 u(x), \  x\in [0,1], \\
        && g(x=0) = g(x=1) = 0, \\
        && u(x) \sim \mathcal{GP}(m(x), k(x,x')),
    \end{aligned}
    \label{eq: 1D SPDE}
\end{equation}
where $u(x)$ is the spatially random forcing function, modelled as a Gaussian process ($\mathcal{GP}$) with mean $m(x)$ set to be 0, and the radial basis function kernel $k(x,x') = \exp \left( -(x-x')^2 / (2l^2) \right)$ where the kernel length scale of the GP is set to be $l=0.1$. The HF dataset is generated using a grid size $\Delta x =0.01$, while the LF dataset grid size is  $\Delta x =0.1$.

In this problem, we consider $2000$ input-output pairs, where the input $u(x)$ and the output $g(x)$ are discretised over $100$ uniformly spaced points in the spatial domain $x$. The dataset is partitioned into $900$ test samples, $100$ validation samples, and $1000$ samples reserved for training. The training set is further randomised to generate subsets with varying sizes of $n_{\text{train}}$.

To construct input intervals $u(x)$, we first sample a random interval length $l_i \sim \mathcal{U}(l_{\min}, l_{\max})$ for each instance $i$. Each $u_i(x)$ is then assigned a fixed interval length $l_i$ across its 100 discretisation points, ensuring uniformity within each input but variability across instances. The corresponding interval-valued outputs $g(x)$ are obtained by propagating these inputs through the true function.  

\begin{figure}[ht]
    \begin{center}
    \includegraphics[width=.72\textwidth]{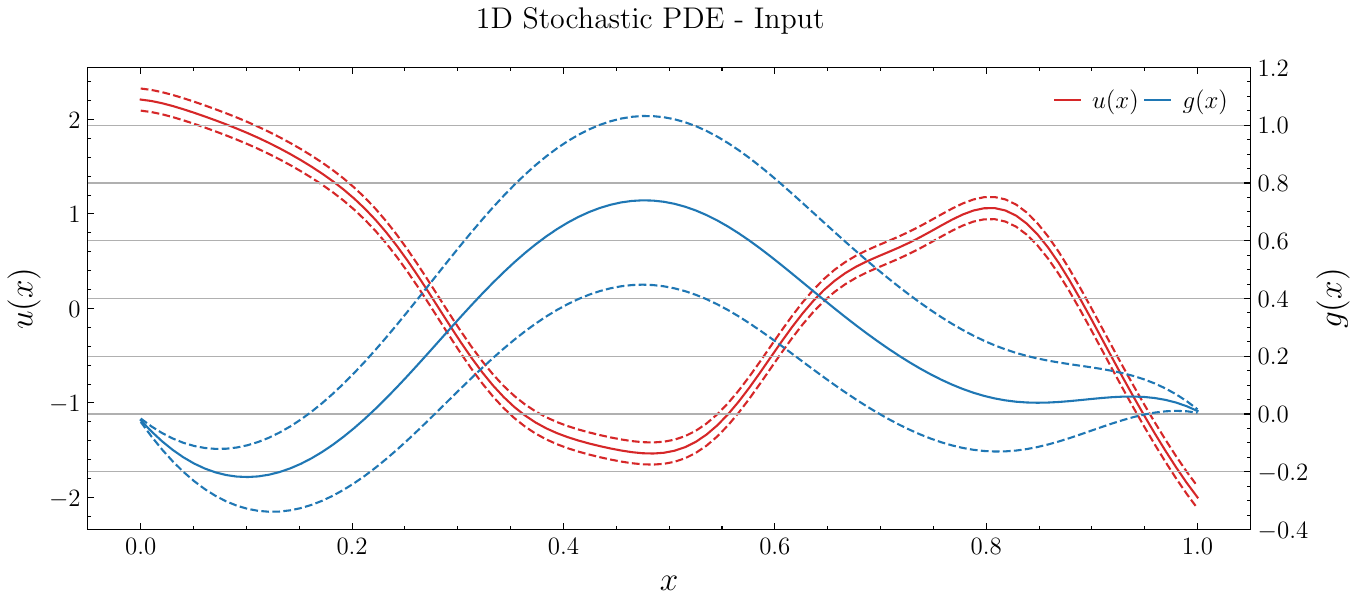}
    \end{center}
\caption{Illustration of an ideal interval dataset for a one-dimensional PDE. The red curve denotes the input function with its interval uncertainty (dashed lines), while the blue curve denotes the corresponding PDE solution with its propagated interval uncertainty.}
\label{fig: ideal_1dPDE}
\end{figure}

Propagating intervals through the optimisation in equation~\ref{eq: interval_prop} is computationally expensive for PDE problems, as both $g_L(x)$ and $g_U(x)$ must be solved at each discretisation point for all samples, resulting in $n_{\text{grid}} \times n_{\text{sample}} \times 2$ optimisation problems. For simplicity, we consider a test case where the lower and upper input bounds $u_L(x)$ and $u_U(x)$ consistently yield $g_L(x)$ and $g_U(x)$, respectively. This assumption, however, does not generally hold for PDEs with non-monotonic behaviour.
An example of a single interval-valued input-output pair for the 1D PDE case is illustrated in Figure~\ref{fig: ideal_1dPDE}.

To construct the DeepONet training dataset, each realisation of the stochastic forcing function $u(x)$ is treated as a single functional input sample paired with its corresponding PDE solution $g(x)$. The input function $u(x)$ is discretised at $m$ sensor locations $\{x_i\}_{i=1}^{m}$, forming a vector representation $\mathbf{u} = [u(x_1), \dots, u(x_m)]$. The output function $g(x)$ is evaluated at $p$ spatial coordinates $\{x_j\}_{j=1}^{p}$, where each coordinate $x_j$ serves as a query location for the solution field. During training, the branch network receives the discretised forcing function $\mathbf{u}$, while the trunk network receives the spatial coordinate $x_j$. The DeepONet then predicts the solution value $g(x_j)$ through the inner product of the latent representations produced by the branch and trunk networks.

The training data are constructed by pairing each functional input $\mathbf{u}$ with all spatial query points $\{x_j\}_{j=1}^{p}$. Consequently, one functional realisation generates multiple training tuples $(\mathbf{u}, x_j, g(x_j))$. The full training dataset is obtained by aggregating these tuples across all sampled GP realisations. During training, mini-batches are formed by randomly sampling these tuples from the aggregated dataset. The network parameters are trained by minimising the mean squared error between the predicted and reference solution values over all sampled tuples. The typical DeepONet architecture for this type of problem consists of relatively shallow fully connected neural networks, usually with 2-4 hidden layers and 32-64 neurons per layer. To select an appropriate architecture, we conduct a brief hyperparameter study focusing on the two best-performing methods. The results are summarised in Table~\ref{tab: ablation_1DPDE}. We eventually chose 4 hidden layers with 64 neurons for our backbone architecture to be applied to all methods.

Analogous to standard regression, the prediction error is decomposed into contributions from the lower and upper bounds. In many engineering applications, conservative prediction intervals are preferred over overly optimistic estimates, as underestimating bounds may lead to unsafe designs or system violations. To reflect this preference, we employ the Linex loss (Eq.~\ref{eq:linex_loss}) with $a=3$, which penalises excessively narrow predictions more strongly. We emphasise that the PDE test case itself does not require conservative interval estimation; rather, the asymmetric loss is introduced to emulate scenarios where conservative predictions are desirable and to evaluate the behaviour of the proposed methods under such assumptions.

The baseline optimisation-based interval propagation method (Opt-Prop) is not considered in this PDE experiment due to the prohibitive computational cost associated with solving an optimisation problem at each discretisation point for every sample. Instead, this study focuses on comparing direct interval propagation approaches. The results presented in Table~\ref{tab: ideal_1dPDE} and Table~\ref{tab: ideal_1dPDE_metrics} show that the CROWN method achieves the lowest Linex error among all evaluated methods, whereas the INN approach exhibits slightly improved coverage performance, as reflected by the CWC metric. This observation suggests that, although INNs provide better interval coverage, they do not capture the shape of the true solution bounds as accurately as the CROWN method.

The relatively poor Linex error observed for the IBP method at training data sizes of 50 and 100 is likely attributable to overly narrow interval predictions compared to the ground-truth bounds, which is consistent with the PINAW values below unity reported in Table~\ref{tab: ideal_1dPDE_metrics}. In contrast, both the Mid-CROWN and Mid-IBP methods consistently generate overly conservative interval estimates, as indicated by their elevated PINAW values, which consequently results in suboptimal Linex error performance. These results highlight the importance of evaluating both coverage-based and error-based metrics when assessing interval prediction methods. A method may achieve low prediction error while providing insufficient coverage, or conversely, attain satisfactory coverage at the expense of accurately capturing the shape of the true bounds, thereby leading to higher error values.

The computational costs of the methods reported in Table~\ref{tab: ideal_1dpde_time} are comparable, with the slowest method requiring approximately one second to evaluate the entire test set. The IBP and CROWN-based implementations are not directly comparable to the Naive and INN approaches, as the former are implemented in \texttt{PyTorch} whereas the latter are implemented in \texttt{TensorFlow}. Although the Opt-Prop method is not explicitly implemented, results from Table~\ref{tab: ideal_1dreg_time} indicate that it is approximately three orders of magnitude slower for a single evaluation. When extended to the PDE setting, where interval propagation must be performed across 100 discretisation points, the computational cost increases proportionally. Consequently, the proposed direct interval prediction methods are estimated to be approximately five orders of magnitude faster than the Opt-Prop baseline.

\begin{table}[!ht]
\centering
\caption{Ideal 1-dimensional PDE linear-exponential (Linex) error comparison across methods and training set sizes.}
{\small
\setlength{\tabcolsep}{4pt}
\begin{tabular}{c c | c c c c c c}
\toprule
\textbf{$n_\text{train}$} & \textbf{Metric} & \textbf{Naive} & \textbf{IBP} & \textbf{CROWN} & \textbf{INN} & \textbf{Mid-CROWN} & \textbf{Mid-IBP} \\
\midrule
\multirow{2}{*}{50}  & $\text{Linex}_L$ & $0.370 \pm 0.125$ & 1529.564 $\pm$ 665.146 & 0.672 $\pm$ 0.690 & 8.059 $\pm$ 10.303 & 4.807 $\pm$ 4.204 & 27.859 $\pm$ 18.670 \\
                     & $\text{Linex}_U$ & $0.348 \pm 0.062$ & 237.603 $\pm$ 101.801 & 9.624 $\pm$ 18.677 & 9.730 $\pm$ 7.674 & 4.863 $\pm$ 4.338 & 27.847 $\pm$ 18.382\\
\hline 
\multirow{2}{*}{100} & $\text{Linex}_L$  & $0.193 \pm 0.054$ & 1493.405 $\pm$ 345.786 & 0.264 $\pm$ 0.213 & 0.728 $\pm$ 1.420 & 2.834 $\pm$ 2.297 & 27.860 $\pm$ 18.211\\
                     & $\text{Linex}_U$  & $0.199 \pm 0.060$ & 247.058 $\pm$ 57.860 & 0.268 $\pm$ 0.222 & 0.842 $\pm$ 1.519 & 2.834 $\pm$ 2.298 & 27.842 $\pm$ 17.782\\
\hline 
\multirow{2}{*}{250} & $\text{Linex}_L$  & $0.077 \pm 0.017$ & 0.681 $\pm$ 1.673 & 0.034 $\pm$ 0.010 & 0.216 $\pm$ 0.295 & 1.127 $\pm$ 1.044 & 17.795 $\pm$ 11.973\\
                     & $\text{Linex}_U$  & $0.078 \pm 0.017$ & 0.657 $\pm$ 1.782 & 0.032 $\pm$ 0.007 & 0.444 $\pm$ 0.542 & 1.121 $\pm$ 1.044 & 17.796 $\pm$ 11.975\\
\hline
\multirow{2}{*}{500} & $\text{Linex}_L$  & $0.042 \pm 0.022$ & 0.683 $\pm$ 1.806 & 0.012 $\pm$ 0.005 & 0.029 $\pm$ 0.050 & 0.749 $\pm$ 0.300 & 9.433 $\pm$ 7.054\\
                     & $\text{Linex}_U$  & $0.043 \pm 0.022$ & 0.648 $\pm$ 1.781 & 0.013 $\pm$ 0.006 & 0.032 $\pm$ 0.057 & 0.749 $\pm$ 0.300 & 9.403 $\pm$ 7.054\\
\bottomrule
\end{tabular}
}
\label{tab: ideal_1dPDE}
\end{table}

\begin{table}[!ht]
\centering
\caption{Ideal 1-dimensional PDE performance comparison across methods and training set sizes on three different metrics: predicted interval normalized average width (PINAW), predictied interval coverage probability (PICP), and the coverage width-based criterion (CWC).}
{\small
\setlength{\tabcolsep}{4pt}
\begin{tabular}{c c | c c c c c c}
\toprule
\textbf{$n_\text{train}$} & \textbf{Metric} & \textbf{Naive} & \textbf{IBP} & \textbf{CROWN} & \textbf{INN} & \textbf{Mid-CROWN} & \textbf{Mid-IBP} \\
\midrule
\multirow{3}{*}{50}  & PINAW & $1.247 \pm 1.004$ & $0.801 \pm 0.299$ & $1.516 \pm 0.810$ & $5.947 \pm 8.095$ & $4.452 \pm 3.606$ & $44.056 \pm 6.963$ \\
                     & PICP & $0.583 \pm 0.224$ & $0.151 \pm 0.203$ & $0.568 \pm 0.274$ & $0.724 \pm 0.260$ & $0.908 \pm 0.171$ & $0.996 \pm 0.042$\\
                     & CWC & $13.097 \pm 18.295$ & $69.769 \pm 44.149$ & $25.230 \pm 29.304$ & $61.983 \pm 236.542$ & $11.717 \pm 10.837$ & $90.025 \pm 27.421$\\
\hline 
\multirow{3}{*}{100} & PINAW  & $1.135 \pm 0.502$ & $0.782 \pm 0.306$ & $1.069 \pm 0.123$ & $1.155 \pm 0.525$ & $2.383 \pm 1.224$ & $55.671 \pm 7.481$ \\
                     & PICP  & $0.730 \pm 0.163$ & $0.148 \pm 0.201$ & $0.674 \pm 0.200$ & $0.813 \pm 0.156$ & $0.912 \pm 0.144$ & $1.000 \pm 0.007$\\
                     & CWC & $7.081 \pm 8.472$ & $68.333 \pm 43.949$ & $10.313 \pm 12.772$ & $6.474 \pm 19.586$ & $6.406 \pm 4.013$ & $111.421 \pm 14.785$\\
\hline 
\multirow{3}{*}{250} & PINAW  & $1.071 \pm 0.175$ & $1.022 \pm 0.052$ & $1.091 \pm 0.074$ & $1.132 \pm 0.323$ & $2.447 \pm 1.204$ & $31.895 \pm 3.396$ \\
                     & PICP  & $0.807 \pm 0.113$ & $0.517 \pm 0.234$ & $0.811 \pm 0.151$ & $0.851 \pm 0.117$ & $0.933 \pm 0.117$ & $1.000 \pm 0.002$\\
                     & CWC & $4.728 \pm 4.836$ & $23.607 \pm 28.606$ & $5.015 \pm 4.800$ & $4.246 \pm 4.517$ & $5.921 \pm 2.650$ & $63.795 \pm 6.771$\\
\hline
\multirow{3}{*}{500} & PINAW  & $1.020 \pm 0.070$ & $2.065 \pm 0.127$ & $1.079 \pm 0.038$ & $1.021 \pm 0.119$ & $2.171 \pm 0.930$ & $12.548 \pm 1.274$  \\
                     & PICP  & $0.885 \pm 0.080$ & $0.686 \pm 0.246$ & $0.902 \pm 0.093$ & $0.931 \pm 0.064$ & $0.965 \pm 0.070$ & $1.000 \pm 0.002$\\
                     & CWC & $3.035 \pm 1.289$ & $25.836 \pm 39.067$ & $3.074 \pm 1.395$ & $2.586 \pm 1.316$ & $4.704 \pm 1.705$ & $25.104 \pm 2.534$ \\
\bottomrule
\end{tabular}
}
\label{tab: ideal_1dPDE_metrics}
\end{table}

\begin{table}[!ht]
    \centering
    \caption{Ideal 1-dimensional PDE time comparison in seconds (s). The IBP and CROWN methods are implemented using \texttt{AutoLiRPA} library in \texttt{PyTorch} framework. Naive and INN methods are impelented in \texttt{TensorFlow} framework due to the availability of existing codebases. The Opt-Prop method were not included in the time comparison due to its infeasibly high computational cost. All experiments were conducted on MacBook Pro with Apple M1 Pro chip.}
    {\footnotesize
    \setlength{\tabcolsep}{4pt}
    \begin{tabular}{c c | c c c c c c}
    \toprule
    \textbf{$n_\text{train}$} & \textbf{Process} & \textbf{Naive} & \textbf{IBP} & \textbf{CROWN} & \textbf{INN} & \textbf{Mid-CROWN} & \textbf{Mid-IBP}
    \\
    \midrule
    \multirow{2}{*}{50}  & Training & $127.201 \pm 2.375$ & $27.907 \pm 0.573$ & $236.503 \pm 1.294$ & $257.835 \pm 5.166$ & $235.662 \pm 0.981$ & $27.865 \pm 0.446$ \\
                        & Inference & $0.972 \pm 0.125$ & $0.051 \pm 0.002$ & $0.502 \pm 0.006$ & $0.458 \pm 0.030$ & $0.506 \pm 0.010$ & $0.052 \pm 0.001$ \\
    \hline 
    \multirow{2}{*}{100}  & Training & $174.037 \pm 8.729$ & $28.411 \pm 0.194$ & $251.899 \pm 1.708$ & $442.901 \pm 1.381$ & $252.166 \pm 2.666$ & $28.886 \pm 0.761$ \\
                        & Inference & $0.974 \pm 0.103$ & $0.050 \pm 0.001$ & $0.490 \pm 0.005$ & $0.457 \pm 0.008$ & $0.499 \pm 0.005$ & $0.072 \pm 0.059$ \\
    \hline 
    \multirow{2}{*}{250} & Training & $276.729 \pm 10.871$ & $47.077 \pm 0.177$ & $508.158 \pm 2.961$ & $755.305 \pm 18.820$ & $517.749 \pm 7.738$ & $47.476 \pm 0.804$ \\
                        & Inference & $1.059 \pm 0.144$ & $0.053 \pm 0.002$ & $0.527 \pm 0.068$ & $0.438 \pm 0.010$ & $0.499 \pm 0.011$ & $0.052 \pm 0.001$ \\
    \hline
    \multirow{2}{*}{500} & Training & $459.397 \pm 8.322$ & $84.156 \pm 0.422$ & $1001.194 \pm 6.547$ & $1406.230 \pm 37.269$ & $1015.253 \pm 10.145$ & $83.775 \pm 1.742$ \\
                        & Inference & $1.092 \pm 0.097$ & $0.052 \pm 0.001$ & $0.494 \pm 0.008$ & $0.475 \pm 0.038$ & $0.497 \pm 0.006$ & $0.051 \pm 0.001$ \\
    \bottomrule
    \end{tabular}
    }
    \label{tab: ideal_1dpde_time}
\end{table}

\subsubsection{2-dimensional PDE}
\label{sec: 2dpde_ideal}

To evaluate the proposed method on a more complex physical system, we consider a two-dimensional partial differential equation (PDE) test case based on the Darcy flow. The Darcy equation describes fluid motion through porous media and is widely used in geotechnical and civil engineering. Following~\cite{Lu2022}, we study this problem in a 2D triangular domain with a notch. The governing equation is given by:
\begin{equation}
\begin{gathered}
    -\nabla \cdot \left(a(x,y)\nabla u(x,y)\right) = f(x,y), \quad (x,y) \in [0,1]^2,\\
    u(x,y)\big|_{\partial D} \sim \mathcal{GP}\!\left(0,\, \mathcal{K}\!\left((x,y),(x',y')\right)\right),
\end{gathered}
\label{eq:Darcy}
\end{equation}
where $a(x,y)\in\mathbb{R}$ denotes the permeability field, $u(x,y)\in\mathbb{R}$ the pressure field, and $f(x,y)\in\mathbb{R}$ the source term.  

Synthetic training and testing datasets are generated by numerically solving~\eqref{eq:Darcy} under varying boundary conditions. The boundary values $u(x,y)\big|_{\partial D}$ are sampled from a zero-mean Gaussian process with a radial basis function (RBF) kernel:
\begin{equation}
    \mathcal{K}\!\left((x,y),(x',y')\right)
    = \exp\!\left[-\frac{(x-x')^2}{2l_x^2} - \frac{(y-y')^2}{2l_y^2}\right],
\end{equation}
where the length-scale parameters are set to $l_x=l_y=0.2$. The permeability and source fields are constant with $a(x,y)=0.1$ and $f(x,y)=-1$, respectively.

\begin{figure}[ht]
    \begin{center}
    \includegraphics[width=.7\textwidth, trim={7cm 0 5cm 0},clip]{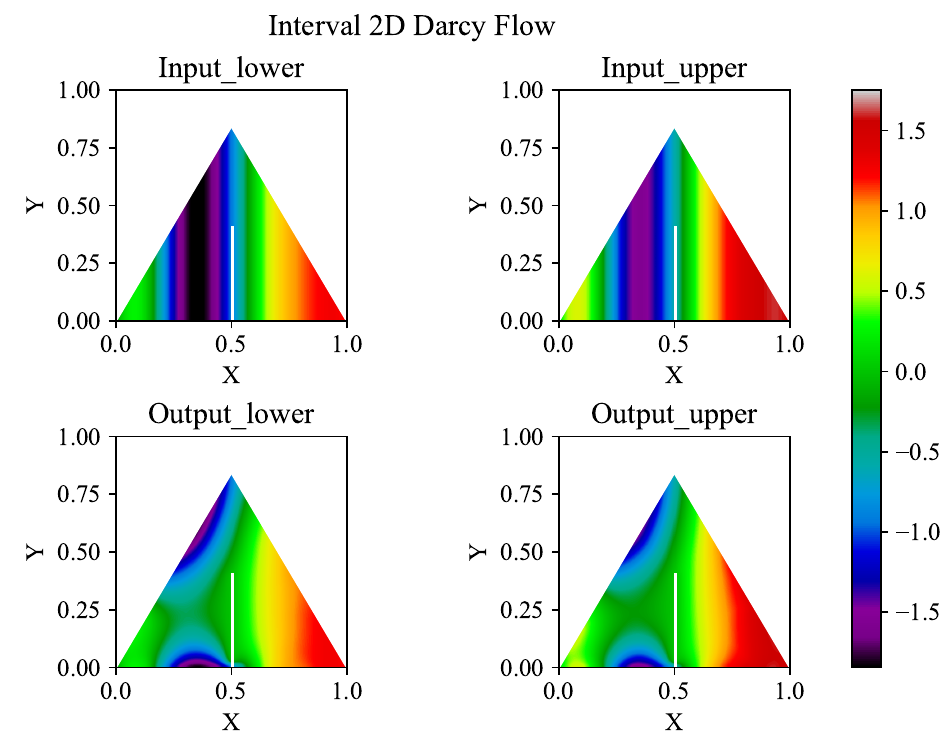}
    \end{center}
\vspace{-0.8cm}
\caption{A single input-output pair of an ideal interval dataset for a one-dimensional PDE.}
\label{fig: interval_2dPDE}
\end{figure}

To construct the interval dataset, we follow a similar procedure from section~\ref{sec: 1dpde_ideal}, where for each input instance $u_i(x)$ in the input set $\boldsymbol{u}(x)$, a fixed interval length $l_i$ is used across its discretisation points. The corresponding interval outputs $g_i(x)$ are obtained by propagating the interval-valued inputs through the true function. Although in this case, similar to section~\ref{sec: 1dpde_ideal}, we choose PDE with monotonic behaviour such that the lower and upper input bounds $u_L(x)$ and $u_U(x)$ consistently yield $g_L(x)$ and $g_U(x)$, for convenience. An example plot for a single instance is illustrated in Figure~\ref{fig: interval_2dPDE}. 

Similar to the one-dimensional PDE case presented in Section~\ref{sec: 1dpde_ideal}, the PDE considered in this test case is steady-state, and therefore the data exhibit no temporal structure. Consequently, the training procedure is identical to that used for the one-dimensional PDE problem. The key difference is that, for the two-dimensional setting, all values defined on the two-dimensional spatial discretisation are flattened before being used as model inputs and outputs. The DeepOnet architecture that is used in this experiment follows the 
setting used in~\cite{Lu2022}

The experimental results in Table~\ref{tab: ideal_2dPDE} indicate that, given sufficient training data, all methods achieve Linex errors of comparable magnitude. However, the coverage metrics reported in Table~\ref{tab: ideal_2dPDE_metrics} reveal notable differences in reliability. In particular, INN struggles to adequately cover the true bounds despite producing relatively wide intervals. In contrast, both Mid-CROWN and Mid-IBP achieve improved CWC performance, even though their Linex errors are slightly higher than that of INN at $n_{\text{train}}=750$. The midpoint-based methods produce larger PINAW values than standard IBP and CROWN because they are not explicitly constrained to learn the interval bounds. Instead, they relax the training objective to approximate only the interval midpoint, which leads to more conservative interval predictions. The inference time in Table~\ref{tab: ideal_2dpde_time} shows that this test case requires more prediction time, although it becomes more apparent that different framework results in different inference time.

\begin{table}[th]
\centering
\caption{Ideal 2-dimensional PDE linear-exponential (Linex) error comparison across methods and training set sizes.}
{\small
\setlength{\tabcolsep}{4pt}
\begin{tabular}{c c | c c c c c c}
\toprule
\textbf{$n_\text{train}$} & \textbf{Metric} & \textbf{Naive} & \textbf{IBP} & \textbf{CROWN} & \textbf{INN} & \textbf{Mid-CROWN} & \textbf{Mid-IBP}\\
\midrule
\multirow{2}{*}{150}  & $\text{Linex}_L$ & $0.148 \pm 0.948$ & $0.070 \pm 0.248$ & $0.056 \pm 0.121$ & $1.447 \pm 22.186$ & $0.127 \pm 0.165$ & $0.165 \pm 0.213$\\
                     & $\text{Linex}_U$ & $0.141 \pm 1.184$ & $0.068 \pm 0.238$ & $0.054 \pm 0.128$ & $1.387 \pm 28.283$ & $0.126 \pm 0.178$ & $0.167 \pm 0.217$\\
\hline 
\multirow{2}{*}{250} & $\text{Linex}_L$  & $0.062 \pm 0.630$ & $0.059 \pm 0.162$ & $0.044 \pm 0.105$ & $0.924 \pm 11.532$ & $0.114 \pm 0.144$ & $0.124 \pm 0.158$\\
                     & $\text{Linex}_U$  & $0.061 \pm 0.687$ & $0.058 \pm 0.157$ & $0.045 \pm 0.104$ & $0.786 \pm 6.351$ & $0.113 \pm 0.150$ & $0.121 \pm 0.177$ \\
\hline
\multirow{2}{*}{500} & $\text{Linex}_L$  & $0.027 \pm 0.398$ & $0.043 \pm 0.102$ & $0.034 \pm 0.057$ & $0.099 \pm 0.489$ & $0.088 \pm 0.113$ & $0.079 \pm 0.098$\\
                     & $\text{Linex}_U$  & $0.027 \pm 0.288$ & $0.042 \pm 0.092$ & $0.034 \pm 0.057$ & $0.095 \pm 0.527$ & $0.085 \pm 0.110$ & $0.080 \pm 0.099$\\
\hline
\multirow{2}{*}{750} & $\text{Linex}_L$  & $0.017 \pm 0.197$ & $0.040 \pm 0.079$ & $0.031 \pm 0.049$ & $0.039 \pm 0.320$ & $0.071 \pm 0.092$ & $0.069 \pm 0.085$\\
                     & $\text{Linex}_U$  & $0.016 \pm 0.135$ & $0.040 \pm 0.075$ & $0.031 \pm 0.050$ & $0.039 \pm 0.293$ & $0.071 \pm 0.091$ & $0.069 \pm 0.087$\\
\bottomrule
\end{tabular}
}
\label{tab: ideal_2dPDE}
\end{table}

\begin{table}[th]
\centering
\caption{Ideal 2-dimensional PDE performance comparison across methods and training set sizes on three different metrics: predicted interval normalized average width (PINAW), predictied interval coverage probability (PICP), and the coverage width-based criterion (CWC).}
{\small
\setlength{\tabcolsep}{4pt}
\begin{tabular}{c c | c c c c c c}
\toprule
\textbf{$n_\text{train}$} & \textbf{Metric} & \textbf{Naive} & \textbf{IBP} & \textbf{CROWN} & \textbf{INN} & \textbf{Mid-CROWN} & \textbf{Mid-IBP} \\
\midrule
\multirow{3}{*}{150}  & PINAW & $1.123 \pm 0.272$ & $1.706 \pm 0.086$ & $1.652 \pm 0.137$ & $2.908 \pm 3.238$ & $2.126 \pm 0.216$ & $2.459 \pm 0.121$ \\
                     & PICP & $0.518 \pm 0.151$ & $0.857 \pm 0.130$ & $0.842 \pm 0.159$ & $0.357 \pm 0.310$ & $0.949 \pm 0.070$ & $0.972 \pm 0.033$\\
                     & CWC & $17.398 \pm 14.332$ & $6.389 \pm 5.329$ & $7.212 \pm 8.503$ & $114.530 \pm 176.641$ & $5.073 \pm 1.778$ & $5.326 \pm 0.591$\\
\hline 
\multirow{3}{*}{250} & PINAW  & $1.060 \pm 0.172$ & $1.679 \pm 0.081$ & $1.574 \pm 0.120$ & $1.833 \pm 1.617$ & $2.087 \pm 0.196$ & $2.221 \pm 0.109$ \\
                     & PICP  & $0.655 \pm 0.133$ & $0.873 \pm 0.121$ & $0.903 \pm 0.109$ & $0.413 \pm 0.292$ & $0.972 \pm 0.037$ & $0.970 \pm 0.043$\\
                     & CWC & $8.684 \pm 7.436$ & $5.772 \pm 4.255$ & $4.710 \pm 3.159$ & $63.371 \pm 102.673$ & $4.523 \pm 0.576$ & $4.874 \pm 0.866$\\
\hline 
\multirow{3}{*}{500} & PINAW  & $1.016 \pm 0.124$ & $1.597 \pm 0.076$ & $1.569 \pm 0.105$ & $1.156 \pm 0.643$ & $1.961 \pm 0.155$ & $1.918 \pm 0.088$ \\
                     & PICP  & $0.735 \pm 0.127$ & $0.922 \pm 0.080$ & $0.946 \pm 0.056$ & $0.589 \pm 0.283$ & $0.980 \pm 0.021$ & $0.982 \pm 0.020$\\
                     & CWC & $5.824 \pm 4.896$ & $4.227 \pm 1.806$ & $3.707 \pm 0.783$ & $22.953 \pm 48.993$ & $4.134 \pm 0.336$ & $4.028 \pm 0.298$\\
\hline
\multirow{3}{*}{750} & PINAW  & $1.044 \pm 0.108$ & $1.555 \pm 0.073$ & $1.522 \pm 0.100$ & $1.105 \pm 0.428$ & $1.849 \pm 0.151$ & $1.847 \pm 0.084$ \\
                     & PICP  & $0.802 \pm 0.095$ & $0.926 \pm 0.075$ & $0.941 \pm 0.063$ & $0.716 \pm 0.231$ & $0.977 \pm 0.029$ & $0.974 \pm 0.026$\\
                     & CWC & $4.261 \pm 2.492$ & $4.031 \pm 1.574$ & $3.688 \pm 0.956$ & $11.458 \pm 30.428$ & $3.935 \pm 0.380$ & $3.965 \pm 0.343$\\
\bottomrule
\end{tabular}
}
\label{tab: ideal_2dPDE_metrics}
\end{table}

\begin{table}[!ht]
    \centering
    \caption{Ideal 2-dimensional PDE time comparison in seconds (s). The IBP and CROWN methods are implemented using \texttt{AutoLiRPA} library in \texttt{PyTorch} framework. Naive and INN methods are impelented in \texttt{TensorFlow} framework due to the availability of existing codebases. The Opt-Prop method were not included in the time comparison due to its infeasibly high computational cost. All experiments were conducted on NVidia P100 GPU.}
    {\footnotesize
    \setlength{\tabcolsep}{4pt}
    \begin{tabular}{c c | c c c c c c}
    \toprule
    \textbf{$n_\text{train}$} & \textbf{Process} & \textbf{Naive} & \textbf{IBP} & \textbf{CROWN} & \textbf{INN} & \textbf{Mid-CROWN} & \textbf{Mid-IBP}
    \\
    \midrule
    \multirow{2}{*}{150}  & Training & $427.811 \pm 5.657$ & $157.825 \pm 1.704$ & $955.687 \pm 21.946$ & $194.951 \pm 5.366$ & $192.697 \pm 3.074$ & $159.386 \pm 1.152$ \\
                        & Inference & $59.407 \pm 2.461$ & $3.972 \pm 0.033$ & $4.966 \pm 0.167$ & $62.753 \pm 2.568$ & $5.086 \pm 0.090$ & $3.938 \pm 0.027$ \\
    \hline 
    \multirow{2}{*}{250}  & Training & $654.959 \pm 8.532$ & $226.241 \pm 4.122$ & $275.351 \pm 8.590$ & $328.865 \pm 3.390$ & $277.282 \pm 5.960$ & $234.361 \pm 5.051$ \\
                        & Inference & $49.773 \pm 0.890$ & $3.944 \pm 0.026$ & $5.216 \pm 0.383$ & $55.475 \pm 2.059$ & $5.326 \pm 0.432$ & $4.025 \pm 0.085$ \\
    \hline 
    \multirow{2}{*}{500} & Training & $1048.473 \pm 44.902$ & $401.825 \pm 9.772$ & $469.127 \pm 18.255$ & $665.211 \pm 14.863$ & $462.445 \pm 13.094$ & $410.800 \pm 8.188$ \\
                        & Inference & $53.315 \pm 9.790$ & $3.996 \pm 0.047$ & $4.859 \pm 0.162$ & $62.299 \pm 8.123$ & $4.868 \pm 0.137$ & $3.930 \pm 0.026$ \\
    \hline
    \multirow{2}{*}{750} & Training & $1562.362 \pm 75.031$ & $562.669 \pm 22.700$ & $651.881 \pm 14.370$ & $996.430 \pm 31.490$ & $654.771 \pm 15.089$ & $558.609 \pm 16.218$ \\
                        & Inference & $45.779 \pm 3.146$ & $3.854 \pm 0.161$ & $4.856 \pm 0.162$ & $56.099 \pm 1.927$ & $4.791 \pm 0.163$ & $4.043 \pm 0.195$ \\
    \bottomrule
    \end{tabular}
    }
    \label{tab: ideal_2dpde_time}
\end{table}

\subsection{Experiments with pointwise data}
\label{sec: exp_aug}

Access to ground-truth interval-valued inputs and outputs is typically unavailable in practical settings; most real-world datasets consist only of pointwise (precise) observations. Therefore, in this experimental setting, we evaluate the robustness and generalisation capability of the models when trained without access to true interval labels, and later evaluated on the actual interval bounds. Consistent with the experiments presented in Section~\ref{sec: exp_ideal}, we perform evaluations on three representative problems: one-dimensional regression, one-dimensional PDE, and two-dimensional PDE.

\subsubsection{Augmentation methods}
\label{sec: aug_methods}
The augmentation procedure aims to transform a pointwise dataset into an interval-valued dataset by leveraging local proximity in the input space $\boldsymbol{x} \in \mathbb{R}^d$. For each proximity-based group, interval bounds are constructed by computing the component-wise minimum and maximum values across both the input vectors $\boldsymbol{x}$ and their associated output targets $\boldsymbol{y}$. This process effectively approximates a local envelope of the input–output mapping. To improve the robustness and generalisation capability of the learned interval model, the grouping procedure is repeated using multiple proximity thresholds. Consequently, the resulting training dataset contains interval samples with heterogeneous widths, representing varying levels of local variability and uncertainty present in the original pointwise data.

To operationalise this strategy, we introduce a grid-based proximity grouping algorithm. The input domain is partitioned into a structured grid along each dimension, where the resolution of the partition is controlled by user-defined grid size parameters. Samples whose input coordinates fall within the same multi-dimensional grid cell are grouped together, and their interval bounds are constructed by computing component-wise extrema over both inputs and outputs within the cell. By performing this grouping procedure across multiple grid resolutions, the algorithm generates interval samples that capture spatial variability at different scales. Coarser grids tend to produce wider intervals that represent broader uncertainty regions, while finer grids generate narrower intervals that capture more localised variations. The full procedure is described in Algorithm~\ref{alg: grid-interval}. 
\begin{algorithm}[ht]
\caption{Grid-based Interval Construction}
\label{alg: grid-interval}
\KwIn{Training data $X$, $Y$; Grid resolutions $R$}
\KwOut{Interval datasets $X_\text{intervals}$, $Y_\text{intervals}$}

$X_\text{intervals} \leftarrow [\emptyset]$, $Y_\text{intervals} \leftarrow [\emptyset]$ \Comment*{Initialize output}

\ForEach{resolution $r \in R$}{
    \Comment{Group data points using grid}
    $Groups \leftarrow$ GridGroup($X$, $r$) \Comment*{Partition data into grid cells}
    
    \ForEach{group $G$ in $Groups$}{
        \Comment{Create intervals from group bounds}
        $X_\text{interval} \leftarrow [\min(X[G]), \max(X[G])]$ \Comment*{Append input interval}
        $Y_\text{interval} \leftarrow [\min(Y[G]), \max(Y[G])]$ \Comment*{Append output interval}
        
    }
}

\Return{$X_\text{intervals}$, $Y_\text{intervals}$}

\end{algorithm}

In high-dimensional settings, grid-based grouping becomes increasingly ineffective due to the curse of dimensionality. For example, even a relatively simple one-dimensional PDE problem discretised over 100 spatial points yields an input space $\boldsymbol{x} \in \mathbb{R}^{100}$. When the available dataset contains only a few hundred samples (e.g., on the order of 500 samples), the sampling density in such a high-dimensional space is extremely low, as the volume of the domain grows exponentially with increasing dimensionality. Under these conditions, uniform grid partitioning produces a large number of cells, most of which remain empty or contain only a small number of samples. As a result, grid-based grouping often fails to produce statistically meaningful interval estimates, since reliable interval construction requires sufficient local sample density to capture input–output variability.

To overcome this limitation, we introduce a clustering-based grouping algorithm that adapts to the intrinsic geometry of the data distribution rather than relying on a predefined spatial partition. In this work, we employ the k-means clustering algorithm \citep{macqueen1967,lloyd1982least} to partition the dataset into groups of samples with similar input characteristics. Unlike grid-based partitioning, clustering methods identify groups based on the proximity between samples in the feature space, allowing the grouping structure to naturally conform to the data manifold. Each cluster is then treated as a local neighbourhood from which interval-valued samples are generated by computing component-wise extrema across the inputs and corresponding outputs of samples belonging to the same cluster. The full procedure is detailed in Algorithm~\ref{alg: clust-interval}.
\begin{algorithm}[ht]
\caption{Clustering-based Interval Construction}
\label{alg: clust-interval}
\KwIn{Training data $X$, $Y$; Number of clusters $K$}
\KwOut{Interval datasets $X_\text{intervals}$, $Y_\text{intervals}$}

$X_\text{intervals} \leftarrow \emptyset$, $Y_\text{intervals} \leftarrow \emptyset$ \Comment*{Initialize output}

\ForEach{cluster count $k \in K$}{
    \Comment{Cluster data points}
    $Labels \leftarrow$ KMeans($X$, $k$) \Comment*{Apply k-means clustering}
    
    \ForEach{cluster $c$ in $\{0, 1, ..., k-1\}$}{
        Find all points $P$ with $Labels[P] = c$ \Comment*{Get points in cluster}
        
        \Comment{Create intervals from cluster bounds}
        $X_{interval} \leftarrow [\min(X[P]), \max(X[P])]$ \Comment*{Append input interval}
        $Y_{interval} \leftarrow [\min(Y[P]), \max(Y[P])]$ \Comment*{Append output interval}
        
    }
}

\Return{$X_\text{intervals}$, $Y_\text{intervals}$}

\end{algorithm}

\subsubsection{1-dimensional regression}
\label{sec: 1dreg_aug}

\begin{figure}[ht]
    \begin{center}
        \begin{subfigure}{.48\textwidth}
            \centering
            \includegraphics[width=.95\textwidth]{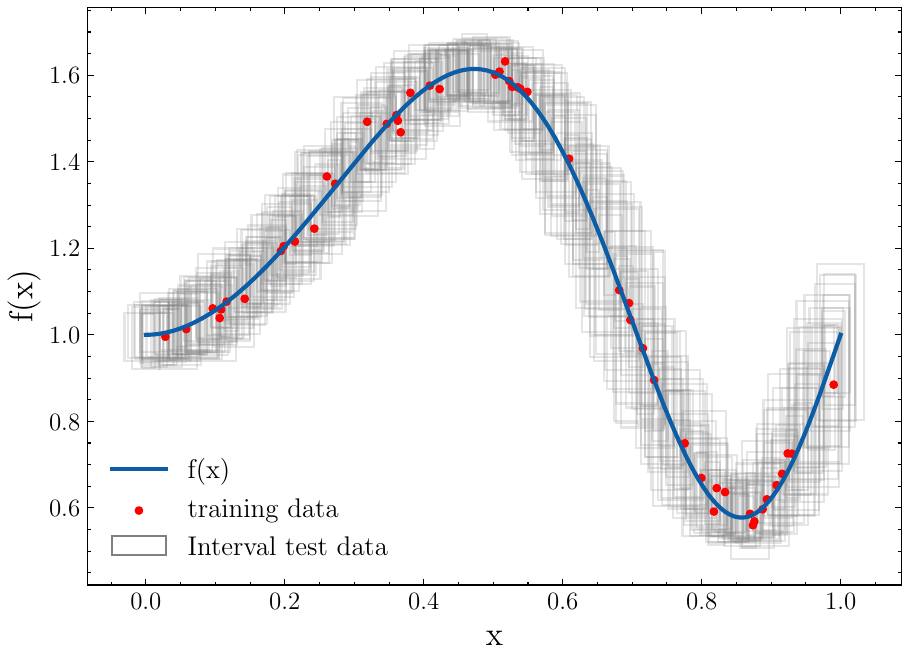}
            \caption{}
            \label{fig: aug_1dreg}
        \end{subfigure}
        \begin{subfigure}{.48\textwidth}
            \centering
            \includegraphics[width=.95\textwidth]{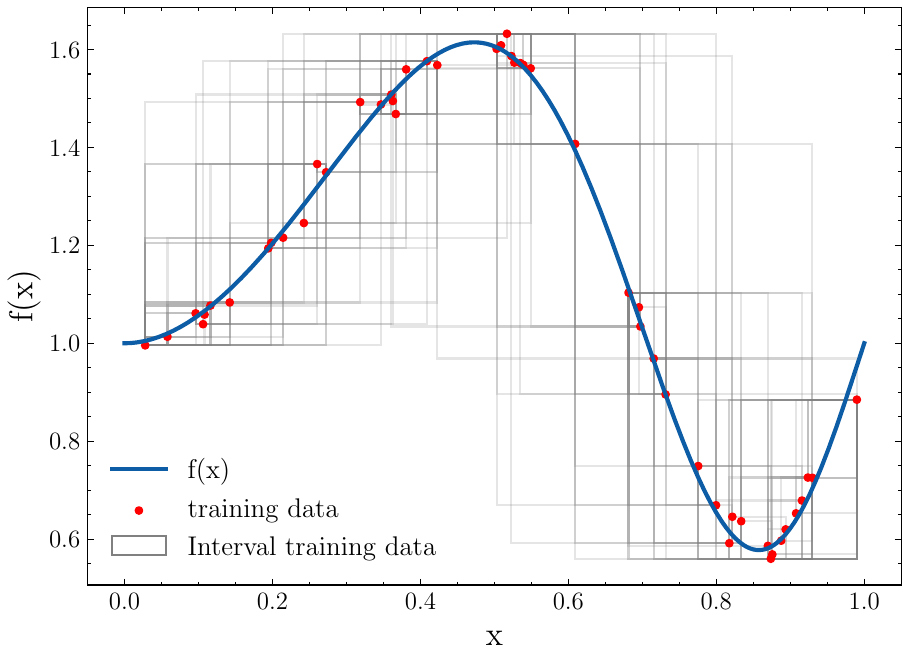}
            \caption{}
            \label{fig: aug_1dreg_train}
        \end{subfigure}
    \end{center}
\vspace{-0.6cm}
\caption{Illustration on (a) Pointwise training dataset with interval test data. (b) Augmented interval training data generated from grid-based interval augmentation.}
\end{figure}

Following the setup described in Section~\ref{sec: 1dreg_ideal}, we consider the same analytical problem defined in Equation~\ref{eq: 1dreg_problem} to evaluate the effectiveness of data augmentation for interval regression. In this setting, rather than having access to a ground-truth interval-valued training set, we assume only pointwise training data is available, while the test set remains interval-valued, as illustrated in Figure~\ref{fig: aug_1dreg}. Given the simplicity of the problem, we adopt the grid-based grouping approach for generating the augmented interval training data.

Figure~\ref{fig: aug_1dreg_train} illustrates the resulting interval-valued training dataset. The figure shows that the augmentation procedure produces intervals of varying widths, resulting from the use of multiple grid resolutions. This variation enables the model to learn from different levels of uncertainty, thereby enhancing its generalisation capability to unseen interval-valued inputs. In this experiment, we construct the interval grouping by using 9 grid resolutions from 0.05 to 0.35, resulting in approximately 150 interval training data. In this experiment, we vary the training set size by randomly sampling from the 150 augmented interval training instances and repeat each configuration ten times with different random seeds to account for stochastic variability. Similar to the ideal interval setting, the direct interval propagation method is compared to the optimisation-based propagation denoted as Opt-Prop.
 
The error metrics, summarised in Table~\ref{tab: aug_1dreg}, indicate that most methods, except IBP and Mid-IBP, achieve accuracy comparable to the Opt-Prop baseline. A closer examination of the interval coverage metrics in Table~\ref{tab: aug_1dPDE_metrics} reveals that most direct interval propagation methods outperform the baseline in terms of coverage quality. Notably, the Mid-IBP method produces significantly larger interval widths, reflected by its elevated PINAW values. Since the employed methods are identical to those used in the ideal interval setting, their training and inference time complexity remains unchanged. The only additional computational overhead arises from the data augmentation step performed prior to training, which is negligible compared to the overall training and inference cost.
 
\begin{table}[!ht]
\centering
\caption{Augmented 1-dimensional regression RMSE comparison across methods and training set sizes.}
{\small
\setlength{\tabcolsep}{4pt}
\begin{tabular}{c c | c c c c c c c}
\toprule
\textbf{$n_\text{train}$} & \textbf{Metric} & \textbf{Naive} & \textbf{IBP} & \textbf{CROWN} & \textbf{INN} & \textbf{Mid-CROWN} & \textbf{Mid-IBP} & \textbf{Opt-Prop}\\
\midrule
\multirow{2}{*}{10}  & $\text{RMSE}_L$ & 0.197 $\pm$ 0.140 & 0.147 $\pm$ 0.075 & 0.118 $\pm$ 0.053 & 0.167 $\pm$ 0.075 & 0.125 $\pm$ 0.095 & 0.458 $\pm$ 0.166 & 0.118 $\pm$ 0.170\\
                      & $\text{RMSE}_U$ & 0.321 $\pm$ 0.195 & 0.170 $\pm$ 0.077 & 0.137 $\pm$ 0.060 & 0.125 $\pm$ 0.055 & 0.149 $\pm$ 0.095 & 0.362 $\pm$ 0.122 & 0.115 $\pm$ 0.141 \\
\hline 
\multirow{2}{*}{25}  & $\text{RMSE}_L$ & 0.065 $\pm$ 0.018 & 0.087 $\pm$ 0.017 & 0.064 $\pm$ 0.011 & 0.062 $\pm$ 0.014 & 0.059 $\pm$ 0.016 & 0.356 $\pm$ 0.115 & 0.052 $\pm$ 0.007 \\
                      & $\text{RMSE}_U$ & 0.082 $\pm$ 0.038 & 0.117 $\pm$ 0.025 & 0.087 $\pm$ 0.035 & 0.067 $\pm$ 0.025 & 0.067 $\pm$ 0.033 & 0.339 $\pm$ 0.115 & 0.047 $\pm$ 0.006 \\
\hline 
\multirow{2}{*}{50} & $\text{RMSE}_L$ & 0.043 $\pm$ 0.009 & 0.091 $\pm$ 0.016 & 0.055 $\pm$ 0.009 & 0.055 $\pm$ 0.009  & 0.047 $\pm$ 0.009 & 0.326 $\pm$ 0.058 & 0.045 $\pm$ 0.007 \\
                      & $\text{RMSE}_U$ & 0.058 $\pm$ 0.015 &  0.119 $\pm$ 0.019 & 0.060 $\pm$ 0.010 & 0.075 $\pm$ 0.038 & 0.044 $\pm$ 0.011 & 0.328 $\pm$ 0.063 & 0.050 $\pm$ 0.006 \\
\hline
\multirow{2}{*}{100} & $\text{RMSE}_L$ & 0.050 $\pm$ 0.007 & 0.085 $\pm$ 0.005 & 0.055 $\pm$ 0.009 & 0.053 $\pm$ 0.016 & 0.046 $\pm$ 0.006 & 0.214 $\pm$ 0.025 & 0.050 $\pm$ 0.005 \\
                      & $\text{RMSE}_U$ & 0.049 $\pm$ 0.009 & 0.102 $\pm$ 0.011 & 0.063 $\pm$ 0.009 & 0.057 $\pm$ 0.012 & 0.045 $\pm$ 0.005 & 0.225 $\pm$ 0.033 & 0.048 $\pm$ 0.005 \\
\bottomrule
\end{tabular}
}
\label{tab: aug_1dreg}
\end{table}

\begin{table}[!ht]
\centering
\caption{Augmented 1-dimensional regression performance comparison across methods and training set sizes on three different metrics: predicted interval normalized average width (PINAW), predictied interval coverage probability (PICP), and the coverage width-based criterion (CWC).}
{\small
\setlength{\tabcolsep}{4pt}
\begin{tabular}{c c | c c c c c c c}
\toprule
\textbf{$n_\text{train}$} & \textbf{Metric} & \textbf{Naive} & \textbf{IBP} & \textbf{CROWN} & \textbf{INN} & \textbf{Mid-CROWN} & \textbf{Mid-IBP} & \textbf{Opt-Prop}\\
\midrule
\multirow{3}{*}{10}  & PINAW & $0.144 \pm 0.940$ & $0.539 \pm 0.071$ & $0.467 \pm 0.037$ & $0.511 \pm 0.624$ & $0.535 \pm 0.061$ & $4.598 \pm 1.417$ & $0.478 \pm 0.102$ \\
                     & PICP & $0.399 \pm 0.086$ & $0.385 \pm 0.077$ & $0.396 \pm 0.031$ & $0.437 \pm 0.134$ & $0.418 \pm 0.101$ & $0.937 \pm 0.042$ & $0.399 \pm 0.053$ \\
                     & CWC & $3.241 \pm 23.543$ & $12.694 \pm 3.873$ & $10.088 \pm 1.088$ & $5.557 \pm 17.497$ & $11.573 \pm 5.795$ & $10.820 \pm 2.684$ & $10.827 \pm 5.546$ \\
\hline 
\multirow{3}{*}{25} & PINAW  & $0.604 \pm 0.223$ & $0.498 \pm 0.032$ & $0.421 \pm 0.025$ & $0.635 \pm 0.138$ & $0.523 \pm 0.041$ & $4.450 \pm 1.332$ & $0.462 \pm 0.018$\\
                     & PICP  & $0.527 \pm 0.081$ & $0.348 \pm 0.062$ & $0.400 \pm 0.030$ & $0.540 \pm 0.077$ & $0.481 \pm 0.064$ & $0.988 \pm 0.014$ & $0.453 \pm 0.016$ \\
                     & CWC & $6.665 \pm 1.226$ & $14.043 \pm 3.788$ & $8.878 \pm 0.776$ & $6.973 \pm 1.361$ & $7.785 \pm 2.276$ & $9.138 \pm 2.606$ & $7.573 \pm 0.367$  \\
\hline 
\multirow{3}{*}{50} & PINAW  & $0.569 \pm 0.055$ & $0.497 \pm 0.033$ & $0.434 \pm 0.012$ & $0.516 \pm 0.173$ & $0.546 \pm 0.013$ & $4.236 \pm 0.657$ & $0.468 \pm 0.015$ \\
                     & PICP  & $0.525 \pm 0.030$ & $0.322 \pm 0.025$ & $0.422 \pm 0.015$ & $0.526 \pm 0.070$ & $0.528 \pm 0.021$ & $0.999 \pm 0.001$ & $0.463 \pm 0.010$  \\
                     & CWC & $6.744 \pm 0.945$ & $15.364 \pm 1.963$ & $8.250 \pm 0.423$ & $5.814 \pm 1.527$ & $6.345 \pm 0.489$ & $8.496 \pm 1.310$ & $7.334 \pm 0.130$\\
\hline
\multirow{3}{*}{100} & PINAW  & $0.560 \pm 0.077$ & $0.473 \pm 0.007$ & $0.427 \pm 0.011$ & $0.541 \pm 0.067$ & $0.541 \pm 0.027$ & $3.180 \pm 0.307$ & $0.446 \pm 0.015$ \\
                     & PICP  & $0.546 \pm 0.057$ & $0.335 \pm 0.028$ & $0.413 \pm 0.014$ & $0.522 \pm 0.066$ & $0.524 \pm 0.025$ & $0.998 \pm 0.002$ & $0.444 \pm 0.013$ \\
                     & CWC & $6.002 \pm 0.709$ & $13.703 \pm 1.724$ & $8.444 \pm 0.352$ & $6.531 \pm 1.180$ & $6.391 \pm 0.450$ & $6.400 \pm 0.620$ & $7.627 \pm 0.228$ \\
\bottomrule
\end{tabular}
}
\label{tab: aug_1dreg_metrics}
\end{table}

\subsubsection{1-dimensional PDE}
\label{sec: 1dpde_aug}

The augmentation of PDE datasets presents unique challenges due to the high dimensionality and spatial correlations inherent in their solutions. As a test case, we consider the one-dimensional second-order stochastic Poisson equation defined in equation~\ref{eq: 1D SPDE}. The PDE is discretised over 100 spatial points, producing an input space $\boldsymbol{x} \in \mathbb{R}^{100}$. To construct interval-valued data, we employ the clustering-based augmentation strategy described in Algorithm~\ref{alg: clust-interval}. Specifically, we generate clusters with sizes varying from 5 to 300, and within each cluster, the minimum and maximum values at each discretization point are used to define interval bounds. This procedure yields approximately 800 interval samples. For the experimental evaluation, we vary the training set size by randomly sampling from these 800 intervals and repeat all experiments 10 times with different random seeds to account for variability.
\begin{figure}[ht]
    \begin{center}
        \begin{subfigure}{.49\textwidth}
            \centering
            \includegraphics[width=1.0\textwidth]{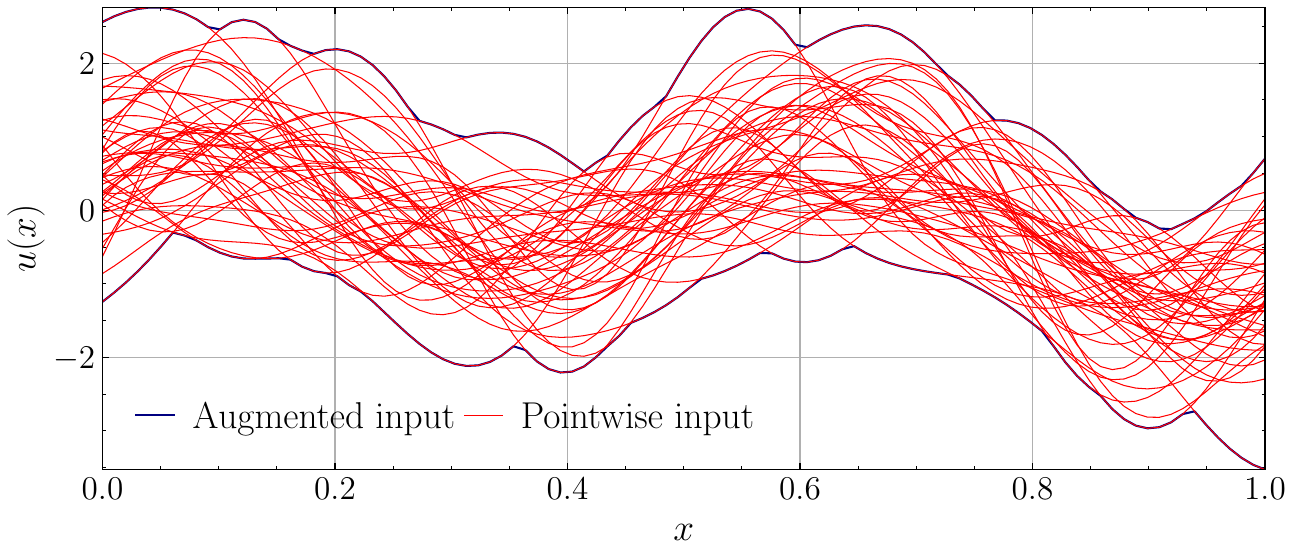}
            \caption{}
            \label{fig: aug_input_1dpde}
        \end{subfigure}
        \begin{subfigure}{.49\textwidth}
            \centering
            \includegraphics[width=1.0\textwidth]{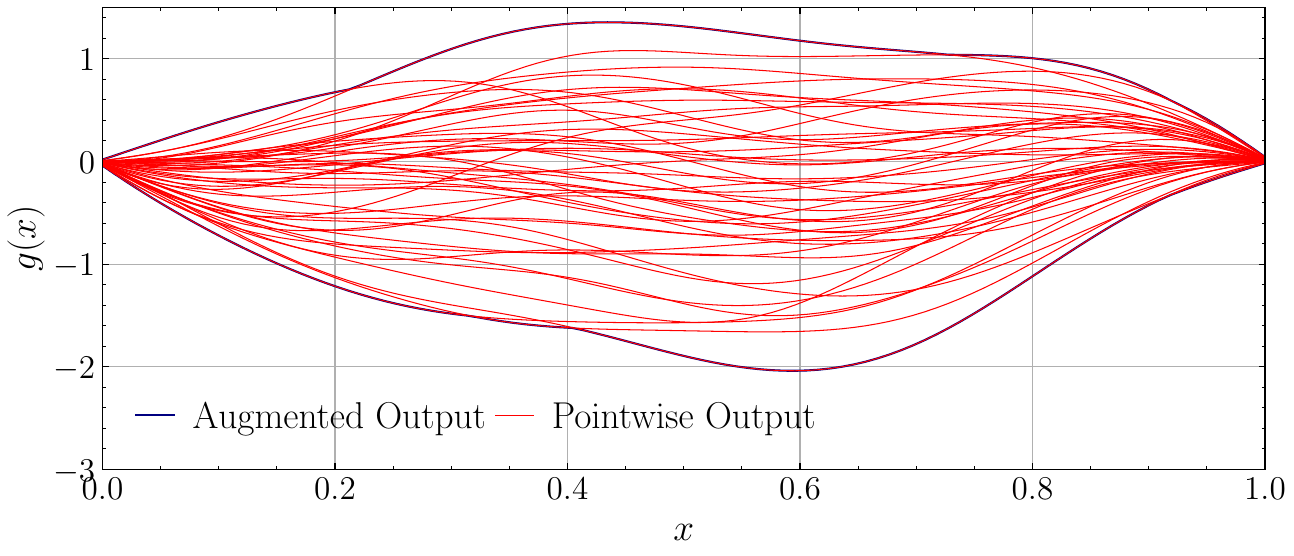}
            \caption{}
            \label{fig: aug_output_1dpde}
        \end{subfigure}
    \end{center}

\vspace{-0.7cm}
\caption{Augmentation result example of a single cluster. (a) Augmented input function. (b) Augmented output function.}
\label{fig: aug_inout_pde}
\end{figure}

\begin{figure}[ht]
    \begin{center}
    \includegraphics[width=.98\textwidth]{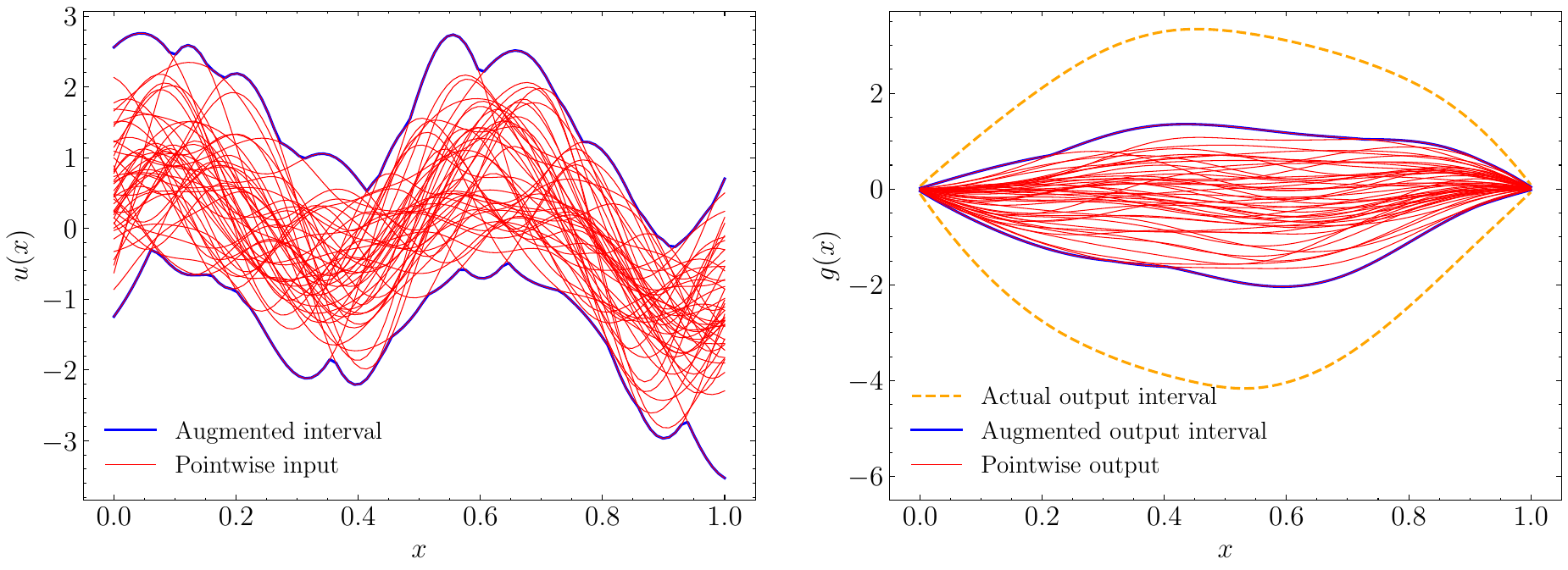}
    \end{center}
\vspace{-0.7cm}
\caption{Augmented input function (left) and comparison between the augmented output and the ground truth output propagated through PDE solver.}
\label{fig: aug_vs_real_1dpde}
\end{figure}

The interval augmentation results, shown in Figure~\ref{fig: aug_inout_pde}, illustrate the constructed input and output intervals. A key challenge lies in the accuracy of these augmented intervals. Specifically, when the augmented input intervals are propagated through a PDE solver, discrepancies arise between the solver's output and the augmented output solved directly from pointwise data (see Figure~\ref{fig: aug_vs_real_1dpde}). While one might consider generating output intervals directly from the PDE solver to avoid such inconsistencies, this would undermine the purpose of employing a surrogate model, as it would still require evaluating a large number of training samples. Consequently, the augmentation process inherently introduces a bias into the model. Nevertheless, the objective is not to perfectly replicate the solver-based intervals but rather to construct interval datasets of sufficient quality to enable effective training of interval-aware models. 

The experimental setup follows the same structure as the ideal interval case described in Section~\ref{sec: 1dpde_ideal}, but the models are trained using augmented interval inputs and outputs. All models are evaluated against the true interval-valued outputs as ground truth. As shown in Table~\ref{tab: aug_1dpde}, the overall performance decreases relative to the ideal scenario in Table~\ref{tab: ideal_1dPDE}, which is expected given the additional uncertainty introduced by data augmentation.

The results presented in Tables~\ref{tab: aug_1dpde} and~\ref{tab: aug_1dPDE_metrics} indicate that Mid-CROWN provides the most balanced performance among the evaluated methods. It achieves low Linex error while maintaining high PICP with relatively tight interval widths, resulting in favourable CWC values. In contrast, the INN and IBP-based methods exhibit comparatively weaker performance, and in several cases the training process fails to converge. We attribute this behaviour to the \emph{interval dependency problem} inherent in interval arithmetic~\cite{Neumaier1993, revol2022affineiterationswrappingeffect}, which can cause excessive overestimation of propagated bounds and lead to optimisation instability. Improving the practical performance of these methods would likely require modifications to the loss formulation or hyperparameter tuning to mitigate bound overestimation. As this study focuses on comparing baseline configurations of each method, such refinements are left for future work.

\begin{table}[!ht]
\centering
\caption{Linear-exponential (Linex) error of augmented 1D PDE experiments across methods and training sizes. For INN and IBP, the asterisk ($^*$) denotes cases where some runs failed to converge or produced \texttt{NaN} values; results are averaged over successful runs only.}
{\small
\setlength{\tabcolsep}{4pt}
\begin{tabular}{c c | c c c c c c}
\toprule
\textbf{$n_\text{train}$} & \textbf{Metric} & \textbf{Naive} & \textbf{IBP}$^*$ & \textbf{CROWN} & \textbf{INN}$^*$ & \textbf{Mid-CROWN} & \textbf{Mid-IBP} \\
\midrule
\multirow{2}{*}{50}  & $\text{Linex}_L$ & $1.559 \pm 0.495$ & 2137.172 $\pm$ 1059.331 & 20.695 $\pm$ 48.603 & 89.456 $\pm$ 92.321 & 0.341 $\pm$ 0.083 & 82.23 $\pm$ 85.715 \\
                      & $\text{Linex}_U$ & $1.610 \pm 0.653$ & 898.399 $\pm$ 442.959 & 40.634 $\pm$ 106.411 & 260.384 $\pm$ 299.820 & 0.390 $\pm$ 0.090 & 42.949 $\pm$ 39.487 \\
\hline 
\multirow{2}{*}{100}  & $\text{Linex}_L$ & $1.191 \pm 0.210$ & 2082.647 $\pm$ 914.356 & 3.815 $\pm$ 1.002 & 22.755 $\pm$ 15.753 & 0.282 $\pm$ 0.075 & 82.573 $\pm$ 98.971 \\
                      & $\text{Linex}_U$ & $1.354 \pm 0.424$ & 856.115 $\pm$ 345.489 & 3.285 $\pm$ 1.435 & 34.114 $\pm$ 31.953 & 0.469 $\pm$ 0.554 & 25.470 $\pm$ 22.488 \\
\hline 
\multirow{2}{*}{250} & $\text{Linex}_L$ & $0.925 \pm 0.149$ & 2139.45 $\pm$ 560.089 & 3.496 $\pm$ 0.764 & 5.860 $\pm$ 2.474 & 0.162 $\pm$ 0.043 & 29.698 $\pm$ 22.862 \\
                      & $\text{Linex}_U$ & $1.125 \pm 0.198$ & 745.528 $\pm$ 184.611 & 3.486 $\pm$ 0.685 & 8.732 $\pm$ 3.769 & 0.168 $\pm$ 0.067 & 10.974 $\pm$ 2.447 \\
\hline
\multirow{2}{*}{500} & $\text{Linex}_L$ & $1.024 \pm 0.097$ & 11.153 $\pm$ 13.783 & 3.223 $\pm$ 0.897 & 33.282 $\pm$ 39.640 & 0.094 $\pm$ 0.015 & 46.682 $\pm$ 58.594 \\
                      & $\text{Linex}_U$ & $1.007 \pm 0.160$ & 8.742 $\pm$ 12.194 & 2.580 $\pm$ 0.728 & 23.928 $\pm$ 13.581 & 0.108 $\pm$ 0.031 & 9.555 $\pm$ 58.555 \\
\bottomrule
\end{tabular}
}
\label{tab: aug_1dpde}
\end{table}

\begin{table}[th]
\centering
\caption{Augmented 1-dimensional PDE performance comparison across methods and training set sizes on three different metrics: predicted interval normalized average width (PINAW), predictied interval coverage probability (PICP), and the coverage width-based criterion (CWC). For INN and IBP, the asterisk ($^*$) denotes cases where some runs failed to converge or produced \texttt{NaN} values; results are averaged over successful runs only.}
{\small
\setlength{\tabcolsep}{4pt}
\begin{tabular}{c c | c c c c c c}
\toprule
\textbf{$n_\text{train}$} & \textbf{Metric} & \textbf{Naive} & \textbf{IBP}$^*$ & \textbf{CROWN} & \textbf{INN}$^*$ & \textbf{Mid-CROWN} & \textbf{Mid-IBP} \\
\midrule
\multirow{3}{*}{50}  & PINAW & $1.231 \pm 0.947$ & $0.362 \pm 0.034$ & $0.762 \pm 0.024$ & $1121.507 \pm 3416.796$ & $1.380 \pm 0.266$ & $3.647 \pm 1.540$ \\
                     & PICP & $0.390 \pm 0.193$ & $0.069 \pm 0.089$ & $0.086 \pm 0.076$ & $1.000 \pm 0.000$ & $0.708 \pm 0.279$ & $0.649 \pm 0.407$ \\
                     & CWC & $37.297 \pm 46.336$ & $41.290 \pm 13.860$ & $78.941 \pm 24.501$ & $2243.015 \pm 6833.591$ & $18.418 \pm 32.571$ & $67.272 \pm 89.325$ \\
\hline 
\multirow{3}{*}{100} & PINAW  & $1.081 \pm 0.924$ & $0.365 \pm 0.018$ & $0.702 \pm 0.025$ & $110.540 \pm 1580.054$ & $1.378 \pm 0.156$ & $3.646 \pm 1.484$\\
                     & PICP  & $0.378 \pm 0.234$ & $0.069 \pm 0.087$ & $0.257 \pm 0.179$ & $0.707 \pm 0.292$ & $0.808 \pm 0.227$ & $0.668 \pm 0.410$ \\
                     & CWC & $39.501 \pm 67.005$ & $41.766 \pm 13.637$ & $41.301 \pm 30.647$ & $318.042 \pm 3165.910$ & $10.119 \pm 19.879$ & $65.564 \pm 93.063$ \\
\hline 
\multirow{3}{*}{250} & PINAW  & $1.184 \pm 1.203$ & $0.349 \pm 0.013$ & $0.785 \pm 0.019$ & $18.882 \pm 23.188$ & $1.135 \pm 0.094$ & $8.522 \pm 2.472$ \\
                     & PICP  & $0.372 \pm 0.259$ & $0.068 \pm 0.090$ & $0.255 \pm 0.169$ & $0.816 \pm 0.201$ & $0.789 \pm 0.201$ & $0.835 \pm 0.308$  \\
                     & CWC & $48.386 \pm 87.066$ & $40.263 \pm 13.449$ & $45.347 \pm 32.783$ & $141.921 \pm 541.130$ & $7.556 \pm 12.817$ & $66.500 \pm 116.931$\\
\hline
\multirow{3}{*}{500} & PINAW  & $0.626 \pm 0.560$ & $0.609 \pm 0.025$ & $0.640 \pm 0.006$ & $106.835 \pm 1394.963$ & $1.191 \pm 0.098$ & $5.189 \pm 1.604$\\
                     & PICP  & $0.286 \pm 0.189$ & $0.230 \pm 0.162$ & $0.249 \pm 0.195$ & $0.863 \pm 0.290$ & $0.856 \pm 0.159$ & $0.797 \pm 0.338$ \\
                     & CWC & $25.232 \pm 31.493$ & $38.272 \pm 25.563$ & $41.634 \pm 32.278$ & $328.899 \pm 2852.497$ & $5.087 \pm 7.984$ & $50.497 \pm 84.956$\\
\bottomrule
\end{tabular}
}
\label{tab: aug_1dPDE_metrics}
\end{table}

\subsubsection{2-dimensional PDE}
\label{sec: 2dpde_aug}

In this section, we consider a two-dimensional Darcy flow PDE with only precise observations available, necessitating interval augmentation. Similar to the one-dimensional case discussed in Section~\ref{sec: 1dpde_aug}, we apply the clustering-based augmentation strategy outlined in Algorithm~\ref{alg: clust-interval}. Due to the higher dimensionality of the problem, $51 \times 51$ grid points resulting in 2601 dimensions, we construct clusters of varying sizes, ranging from 5 to 300. This procedure yields approximately 860 augmented interval samples in total. 

The results presented in Tables~\ref{tab: aug_2dPDE} and~\ref{tab: aug_2dPDE_metrics} indicate that Mid-IBP achieves the best overall performance, as reflected by both the Linex error and interval coverage metrics. This observation differs from the augmented one-dimensional PDE test case in Section~\ref{sec: 1dpde_aug}, where Mid-IBP showed less competitive performance. While the precise reason for this discrepancy is unclear, it suggests that the relative performance of the interval propagation methods may depend on characteristics of the underlying PDE and data distribution rather than dimensionality alone. For instance, the two-dimensional Darcy flow problem may present structural properties that are more amenable to learning by midpoint-based approaches, although further investigation would be required to confirm this hypothesis. The INN method remains the lowest-performing approach in this test case. However, its performance shows improvement compared to the one-dimensional PDE results. 

\begin{table}[th]
\centering
\caption{Augmented 2-dimensional PDE linear-exponential (Linex) error comparison across methods and training set sizes.}
{\small
\setlength{\tabcolsep}{4pt}
\begin{tabular}{c c | c c c c c c}
\toprule
\textbf{$n_\text{train}$} & \textbf{Metric} & \textbf{Naive} & \textbf{IBP} & \textbf{CROWN} & \textbf{INN} & \textbf{Mid-CROWN} & \textbf{Mid-IBP} \\
\midrule
\multirow{2}{*}{150}  & $\text{Linex}_L$ & 0.224 $\pm$ 1.946 & 0.207 $\pm$ 0.590 & 0.140 $\pm$ 0.857 & 12.876 $\pm$ 1.171 & 0.086 $\pm$ 0.018 & 0.075 $\pm$ 0.021\\
                     & $\text{Linex}_U$ & 0.212 $\pm$ 1.076 & 0.192 $\pm$ 0.526 & 0.131 $\pm$ 0.616 & 9.832 $\pm$ 0.868 & 0.089 $\pm$ 0.032 & 0.081 $\pm$ 0.011\\
\hline 
\multirow{2}{*}{250} & $\text{Linex}_L$  & 0.119 $\pm$ 0.063 & 0.175 $\pm$ 0.049 & 0.102 $\pm$ 0.016 & 14.413 $\pm$ 1.214 & 0.074 $\pm$ 0.003 & 0.060 $\pm$ 0.005\\
                     & $\text{Linex}_U$  & 0.111 $\pm$ 0.078 & 0.166 $\pm$ 0.048 & 0.086 $\pm$ 0.010 & 9.022 $\pm$ 1.201 & 0.078 $\pm$ 0.008 & 0.059 $\pm$ 0.005\\
\hline 
\multirow{2}{*}{500} & $\text{Linex}_L$  & 0.066 $\pm$ 0.003 & 0.103 $\pm$ 0.021 & 0.067 $\pm$ 0.007 & 1.728 $\pm$ 0.651 & 0.065 $\pm$ 0.007 & 0.042 $\pm$ 0.003\\
                     & $\text{Linex}_U$  & 0.064 $\pm$ 0.003 & 0.100 $\pm$ 0.022 & 0.060 $\pm$ 0.006 & 1.317 $\pm$ 0.661 & 0.069 $\pm$ 0.005 & 0.044 $\pm$ 0.003\\
\hline
\multirow{2}{*}{750} & $\text{Linex}_L$  & 0.047 $\pm$ 0.002 & 0.095 $\pm$ 0.015 & 0.062 $\pm$ 0.006 & 0.310 $\pm$ 0.100 & 0.060 $\pm$ 0.003 & 0.037 $\pm$ 0.065\\
                     & $\text{Linex}_U$  & 0.047 $\pm$ 0.002 & 0.092 $\pm$ 0.015 & 0.055 $\pm$ 0.003 & 0.289 $\pm$ 0.104 & 0.066 $\pm$ 0.002 & 0.041 $\pm$ 0.007\\
\bottomrule
\end{tabular}
}
\label{tab: aug_2dPDE}
\end{table}

\begin{table}[th]
\centering
\caption{Augmented 2-dimensional PDE performance comparison across methods and training set sizes on three different metrics: predicted interval normalized average width (PINAW), predictied interval coverage probability (PICP), and the coverage width-based criterion (CWC).}
{\small
\setlength{\tabcolsep}{4pt}
\begin{tabular}{c c | c c c c c c}
\toprule
\textbf{$n_\text{train}$} & \textbf{Metric} & \textbf{Naive} & \textbf{IBP} & \textbf{CROWN} & \textbf{INN} & \textbf{Mid-CROWN} & \textbf{Mid-IBP} \\
\midrule
\multirow{3}{*}{150}  & PINAW & $1.671 \pm 0.563$ & $1.488 \pm 0.072$ & $1.536 \pm 0.130$ & $3.604 \pm 3.465$ & $1.911 \pm 0.194$ & $1.793 \pm 0.090$ \\
                     & PICP & $0.612 \pm 0.177$ & $0.734 \pm 0.161$ & $0.737 \pm 0.206$ & $0.285 \pm 0.192$ & $0.911 \pm 0.105$ & $0.909 \pm 0.091$\\
                     & CWC & $20.748 \pm 26.414$ & $9.886 \pm 10.313$ & $11.946 \pm 14.837$ & $134.387 \pm 116.458$ & $5.429 \pm 2.913$ & $5.024 \pm 2.248$\\
\hline 
\multirow{3}{*}{250} & PINAW  & $1.558 \pm 0.430$ & $1.519 \pm 0.073$ & $1.511 \pm 0.095$ & $3.157 \pm 2.823$ & $1.879 \pm 0.199$ & $1.733 \pm 0.083$ \\
                     & PICP  & $0.710 \pm 0.157$ & $0.636 \pm 0.197$ & $0.818 \pm 0.159$ & $0.266 \pm 0.180$ & $0.907 \pm 0.110$ & $0.911 \pm 0.105$\\
                     & CWC & $11.877 \pm 14.653$ & $17.284 \pm 19.516$ & $7.037 \pm 7.087$ & $133.544 \pm 108.901$ & $5.510 \pm 3.513$ & $5.116 \pm 4.350$\\
\hline 
\multirow{3}{*}{500} & PINAW  & $1.364 \pm 0.309$ & $1.459 \pm 0.068$ & $1.473 \pm 0.081$ & $1.712 \pm 1.793$ & $1.777 \pm 0.179$ & $1.659 \pm 0.078$ \\
                     & PICP  & $0.750 \pm 0.167$ & $0.839 \pm 0.126$ & $0.845 \pm 0.152$ & $0.261 \pm 0.206$ & $0.936 \pm 0.080$ & $0.949 \pm 0.055$\\
                     & CWC & $9.307 \pm 15.618$ & $5.717 \pm 4.596$ & $6.145 \pm 6.185$ & $92.086 \pm 122.810$ & $4.453 \pm 1.483$ & $3.899 \pm 0.826$\\
\hline
\multirow{3}{*}{750} & PINAW  & $1.411 \pm 0.279$ & $1.505 \pm 0.073$ & $1.461 \pm 0.078$ & $1.450 \pm 1.779$ & $1.826 \pm 0.194$ & $1.642 \pm 0.077$ \\
                     & PICP  & $0.820 \pm 0.127$ & $0.770 \pm 0.160$ & $0.845 \pm 0.151$ & $0.368 \pm 0.258$ & $0.930 \pm 0.082$ & $0.950 \pm 0.055$\\
                     & CWC & $6.093 \pm 6.228$ & $8.631 \pm 9.119$ & $6.052 \pm 6.264$ & $67.788 \pm 111.043$ & $4.659 \pm 1.556$ & $3.852 \pm 0.868$\\
\bottomrule
\end{tabular}
}
\label{tab: aug_2dPDE_metrics}
\end{table}

\section{Discussion}
\label{sec: discussion}
In this work, we conducted a comprehensive investigation of interval uncertainty propagation in neural network-based surrogate models, focusing on both multilayer perceptrons (MLPs) and Deep Operator Networks (DeepONets). We examined three classes of approaches: (i) a naive interval propagation method, (ii) bound propagation techniques including IBP and CROWN, and (iii) interval neural networks (INNs) employing interval-valued weights. The methods were evaluated under two training scenarios. The first assumes access to interval-valued datasets containing paired interval inputs and outputs, which are often scarce and costly to obtain. The second considers the more practical setting where only pointwise training data are available. The evaluation was performed on three representative test cases: a one-dimensional regression problem as a proof of concept, a one-dimensional PDE, and a two-dimensional PDE.

We further compared the direct interval propagation approaches against the standard optimisation-based propagation baseline (Opt-Prop). In the one-dimensional regression experiments, under both idealised and pointwise training conditions, several direct interval propagation methods achieved comparable or superior performance in terms of accuracy and interval coverage metrics. Moreover, the inference cost of direct interval propagation was substantially lower, showing improvements of approximately three orders of magnitude relative to Opt-Prop. The optimisation-based approach also becomes impractical for full-field PDE problems due to the need to solve optimisation problems at each discretisation point. These results suggest that direct interval propagation provides a more computationally efficient and scalable alternative for interval propagation of full-field simulations.

Under ideal interval training conditions, the results indicate that simple architectures, such as the naive approach, can perform remarkably well. Although INNs and bound propagation methods, including IBP, CROWN, and their midpoint variants, outperform the naive approach in certain metrics and test cases, the naive method demonstrates consistently strong overall performance. During training, the lower and upper output nodes in the naive architecture remain coupled through shared hidden layers, allowing flexible modelling of dependencies between interval bounds. This structural flexibility enables the naive approach to effectively exploit the availability of ideal interval-valued training data.

Since ideal interval-valued datasets are rarely available in practice, a more realistic setting is to train interval models using standard pointwise data. To facilitate this, we employ interval augmentation techniques, specifically the grid-based and clustering-based strategies described in Algorithms~\ref{alg: grid-interval} and~\ref{alg: clust-interval}. The results for the augmented one-dimensional regression problem indicate that transforming pointwise data into interval form is effective: strictly constrained models exhibit only a modest performance degradation relative to the ideal interval setting.

In contrast, interval augmentation for PDE problems is substantially more challenging. First, each discretisation point contributes to the overall input dimensionality, such that a PDE discretised on 100 grid points corresponds to a 100-dimensional input space. Second, the augmented intervals do not necessarily represent accurate bounds of the true solution. As illustrated in Figure~\ref{fig: aug_vs_real_1dpde}, noticeable discrepancies arise between the augmented intervals and the reference bounds obtained via interval propagation through the PDE solver, particularly in the one-dimensional PDE case. As a consequence, Table~\ref{tab: aug_1dpde} shows that models trained using direct bound matching on augmented 1D PDE data experience a significant degradation in performance compared to the ideal interval scenario.

In the augmented PDE experiments, the Mid-CROWN model consistently outperforms the naive approach. CROWN-based methods approximate nonlinear activation functions using upper and lower affine relaxations. This linearisation introduces an approximation error, which can reduce prediction accuracy compared with the naive method, particularly when trained on ideal interval-valued data where strict bound matching is possible. Nevertheless, CROWN-based methods demonstrate strong generalisation capability, which explains their consistently competitive performance across all test cases. When training is performed using augmented interval data derived from pointwise observations, directly fitting the augmented bounds can degrade performance, since the augmented intervals do not necessarily represent accurate solution bounds. Relaxing this constraint by training on interval midpoints, while allowing the model to infer appropriate bound widths, appears to improve robustness, which explains the strong performance of Mid-CROWN in the augmented PDE setting.

In contrast, the performance of INN and IBP-based methods exhibits higher variability across experiments. These approaches rely heavily on interval arithmetic and are therefore susceptible to the interval dependency and wrapping effects~\cite{Neumaier1993, revol2022affineiterationswrappingeffect}. These phenomena can lead to bound overestimation and increased optimisation difficulty during training, resulting in inconsistent performance across test cases. However, in certain scenarios, both methods achieve competitive or superior results, indicating their potential effectiveness when appropriately configured in certain cases.  

Additionally, we conduct a brief hyperparameter study on the neural network architecture, as presented in Tables~\ref{tab: ablation_1Dreg} and~\ref{tab: ablation_1DPDE}. The results confirm that the architectural choice influences the prediction performance. However, the variability induced by different architectures is considerably smaller than the variability resulting from the choice of interval propagation strategy.  

\section{Summary and Future Works}

In summary, we have demonstrated that direct interval propagation offers a more effective and computationally efficient alternative to standard optimisation-based interval propagation, particularly for full-field problems such as PDE problems. By reformulating interval uncertainty propagation as an interval-valued regression task, the proposed approaches eliminate the need for repeated surrogate evaluations within inner optimisation loops, leading to substantial computational savings and improved scalability. This enhanced scalability enables more efficient downstream analyses that explicitly account for uncertainty, such as robust and reliability-based optimisation. Moreover, direct interval propagation facilitates the use of full-field PDE surrogate models in interval-based uncertainty analysis, allowing engineers to obtain spatially resolved uncertainty information rather than relying solely on scalar summary quantities.

Through extensive numerical experiments across regression, one-dimensional PDE, and two-dimensional PDE test cases, we showed that direct interval propagation methods can achieve accuracy and interval coverage comparable to, and in some cases exceeding, optimisation-based propagation, while being several orders of magnitude faster at inference time. Among the methods studied, simple architectures such as the naive approach performed remarkably well under ideal interval training conditions, whereas bound propagation techniques, particularly midpoint-based variants such as Mid-CROWN, exhibited improved robustness when trained on augmented interval data derived from pointwise observations. Our results also highlight important trade-offs among different interval propagation strategies. While INN and IBP-based methods can offer strong coverage in certain settings, their performance is more sensitive to interval dependency and wrapping effects inherent to interval arithmetic, leading to greater variability across test cases.

Several directions for future research emerge from this work. A natural extension is the application of direct interval propagation to realistic test cases such as optimisation problems~\cite{Ma2025, Zhao2025} or identification problems~\cite{Zhang2025}, where both computational efficiency and accurate uncertainty representation are critical. Such studies would further demonstrate the practical advantages of direct interval propagation in engineering decision-making tasks.
Another promising direction is the integration of statistical calibration techniques to improve interval coverage and efficiency. The present study primarily relies on deterministic surrogate models, which themselves may introduce modelling uncertainty. Incorporating statistical post-processing methods, such as conformal prediction, could provide calibrated interval estimates with improved reliability guarantees. Combining deterministic interval propagation with statistical calibration may therefore enhance the robustness and trustworthiness of interval predictions in practical applications.

\section*{Data Availability}
The data and code are available at: \url{https://github.com/fazaghifari/IntervalOperatorLearning}. Although, the provided example in the repository are not the exact experiments presented in this paper, they illustrate the implementation of different interval propagation methods discussed herein.

\section*{Acknowledgments}
This work was supported by the granted H2020 FETOPEN-2018-2019-2020-01 European project, \emph{Epistemic AI} under grant agreement No. 964505 (E-pi). The authors acknowledge the use of AI tools, such as ChatGPT and Grammarly, to assist with grammar refinement and improve the clarity of wording in the manuscript. All substantive content, analysis, and conclusions are the original work of the authors.

\newpage
\appendix
\section{Hyperparameter Study}

\begin{table}[!ht]
\centering
\caption{Brief hyperparameter study of neural network for the 1D regression, focusing on the two best-performing methods in each experimental setting on $n_\text{train} = 50$. The configuration ``2L-8'' notation denotes that both the branch and trunk networks consist of 2 hidden layers with 8 neurons per layer.}
{\small
\setlength{\tabcolsep}{4pt}
\begin{tabular}{c c | cc | cc}
\toprule
\multirow{2}{*}{\textbf{Configuration}} & \multirow{2}{*}{\textbf{Metric}} 
& \multicolumn{2}{c|}{\textbf{Ideal}} 
& \multicolumn{2}{c}{\textbf{Augmented}} \\
\cmidrule(lr){3-4} \cmidrule(lr){5-6}
& & \textbf{Naive} & \textbf{INN} & \textbf{Naive} & \textbf{Mid-CROWN} \\
\midrule

\multirow{2}{*}{2L-8}
& $\text{RMSE}_L$ & $0.050 \pm 0.057$ & $0.043 \pm 0.041$ & $0.065 \pm 0.023$ & $0.087 \pm 0.028$ \\
& $\text{RMSE}_U$ & $0.058 \pm 0.062$ & $0.040 \pm 0.044$ & $0.067 \pm 0.032$ & $0.104 \pm 0.043$ \\
\midrule

\multirow{2}{*}{2L-16}
& $\text{RMSE}_L$ & $0.034 \pm 0.041$ & $0.043 \pm 0.039$ & $0.062 \pm 0.036$ & $0.069 \pm 0.017$ \\
& $\text{RMSE}_U$ & $0.047 \pm 0.053$ & $0.034 \pm 0.035$ & $0.075 \pm 0.040$ & $0.077 \pm 0.030$ \\
\midrule

\multirow{2}{*}{2L-32}
& $\text{RMSE}_L$ & $0.021 \pm 0.006$ & $0.031 \pm 0.019$ & $0.046 \pm 0.005$ & $0.063 \pm 0.007$ \\
& $\text{RMSE}_U$ & $0.031 \pm 0.015$ & $0.030 \pm 0.021$ & $0.058 \pm 0.023$ & $0.054 \pm 0.009$ \\
\midrule

\multirow{2}{*}{3L-8}
& $\text{RMSE}_L$ & $0.022 \pm 0.004$ & $0.035 \pm 0.042$ & $0.056 \pm 0.017$ & $0.047 \pm 0.008$ \\
& $\text{RMSE}_U$ & $0.025 \pm 0.011$ & $0.041 \pm 0.043$ & $0.061 \pm 0.030$ & $0.053 \pm 0.010$ \\
\midrule

\multirow{2}{*}{3L-16}
& $\text{RMSE}_L$ & $0.023 \pm 0.005$ & $0.019 \pm 0.004$ & $0.043 \pm 0.009$ & $0.047 \pm 0.009$ \\
& $\text{RMSE}_U$ & $0.026 \pm 0.006$ & $0.019 \pm 0.003$ & $0.058 \pm 0.015$ & $0.044 \pm 0.011$ \\
\midrule

\multirow{2}{*}{3L-32}
& $\text{RMSE}_L$ & $0.019 \pm 0.002$ & $0.024 \pm 0.008$ & $0.045 \pm 0.005$ & $0.043 \pm 0.008$ \\
& $\text{RMSE}_U$ & $0.020 \pm 0.004$ & $0.021 \pm 0.005$ & $0.052 \pm 0.014$ & $0.045 \pm 0.008$ \\

\bottomrule
\end{tabular}
}
\label{tab: ablation_1Dreg}
\end{table}

\begin{table}[!ht]
\centering
\caption{Brief hyperparameter study of DeepONet for the 1D PDE, focusing on the two best-performing methods in each experimental setting on $n_\text{train} = 500$. The configuration ``2L-32'' notation denotes that both the branch and trunk networks consist of 2 hidden layers with 32 neurons per layer.}
{\small
\setlength{\tabcolsep}{4pt}
\begin{tabular}{c c | cc | cc}
\toprule
\multirow{2}{*}{\textbf{Configuration}} & \multirow{2}{*}{\textbf{Metric}} 
& \multicolumn{2}{c|}{\textbf{Ideal}} 
& \multicolumn{2}{c}{\textbf{Augmented}} \\
\cmidrule(lr){3-4} \cmidrule(lr){5-6}
& & \textbf{Naive} & \textbf{CROWN} & \textbf{Naive} & \textbf{Mid-CROWN} \\
\midrule

\multirow{2}{*}{2L-32}
& $\text{Linex}_L$ & $0.108 \pm 0.045$ & $0.019 \pm 0.005$ & $0.912 \pm 0.056$ & $1.749 \pm 0.235$ \\
& $\text{Linex}_U$ & $0.107 \pm 0.045$ & $0.019 \pm 0.004$ & $0.798 \pm 0.069$ & $1.704 \pm 0.243$ \\
\midrule

\multirow{2}{*}{2L-64}
& $\text{Linex}_L$ & $0.152 \pm 0.072$ & $0.012 \pm 0.005$ & $0.904 \pm 0.070$ & $1.223 \pm 0.194$ \\
& $\text{Linex}_U$ & $0.154 \pm 0.073$ & $0.011 \pm 0.005$ & $0.839 \pm 0.095$ & $1.179 \pm 0.228$ \\
\midrule

\multirow{2}{*}{2L-128}
& $\text{Linex}_L$ & $0.173 \pm 0.082$ & $0.009 \pm 0.002$ & $0.946 \pm 0.098$ & $0.713 \pm 0.099$ \\
& $\text{Linex}_U$ & $0.173 \pm 0.082$ & $0.009 \pm 0.002$ & $0.931 \pm 0.116$ & $0.684 \pm 0.155$ \\
\midrule

\multirow{2}{*}{4L-32}
& $\text{Linex}_L$ & $0.045 \pm 0.022$ & $0.023 \pm 0.011$ & $0.970 \pm 0.090$ & $0.114 \pm 0.057$ \\
& $\text{Linex}_U$ & $0.045 \pm 0.023$ & $0.022 \pm 0.008$ & $0.915 \pm 0.071$ & $0.118 \pm 0.055$ \\
\midrule

\multirow{2}{*}{4L-64}
& $\text{Linex}_L$ & $0.042 \pm 0.022$ & $0.012 \pm 0.005$ & $1.024 \pm 0.097$ & $0.094 \pm 0.015$ \\
& $\text{Linex}_U$ & $0.043 \pm 0.022$ & $0.013 \pm 0.006$ & $1.007 \pm 0.160$ & $0.108 \pm 0.031$ \\
\midrule

\multirow{2}{*}{4L-128}
& $\text{Linex}_L$ & $0.023 \pm 0.015$ & $0.033 \pm 0.024$ & $0.699 \pm 0.083$ & $0.073 \pm 0.014$ \\
& $\text{Linex}_U$ & $0.023 \pm 0.015$ & $0.034 \pm 0.023$ & $0.849 \pm 0.122$ & $0.072 \pm 0.017$ \\

\bottomrule
\end{tabular}
}
\label{tab: ablation_1DPDE}
\end{table}

\bibliographystyle{elsarticle-num} 
\bibliography{main-bib}

@incollection{Billard2000,
  doi = {10.1007/978-3-642-59789-3_58},
  url = {https://doi.org/10.1007/978-3-642-59789-3_58},
  year = {2000},
  publisher = {Springer Berlin Heidelberg},
  pages = {369--374},
  author = {L. Billard and E. Diday},
  title = {Regression Analysis for Interval-Valued Data},
  booktitle = {Studies in Classification,  Data Analysis,  and Knowledge Organization}
}

@article{LimaNeto2010,
  doi = {10.1016/j.csda.2009.08.010},
  url = {https://doi.org/10.1016/j.csda.2009.08.010},
  year = {2010},
  month = feb,
  publisher = {Elsevier {BV}},
  volume = {54},
  number = {2},
  pages = {333--347},
  author = {Eufr{\'{a}}sio de A. Lima Neto and Francisco de A.T. de Carvalho},
  title = {Constrained linear regression models for symbolic interval-valued variables},
  journal = {Computational Statistics and Data Analysis}
}

@inproceedings{Faza2024,
  title = {Interval Reduced Order Surrogate Modelling Framework for Uncertainty Quantification},
  url = {http://dx.doi.org/10.2514/6.2024-0387},
  DOI = {10.2514/6.2024-0387},
  booktitle = {AIAA SCITECH 2024 Forum},
  pages = {0},
  publisher = {American Institute of Aeronautics and Astronautics},
  author = {Faza,  Ghifari A. and Shariatmadar,  Keivan and Hallez,  Hans and Moens,  David},
  year = {2024},
  month = jan 
}

@article{Roque2007,
  title = {iMLP: Applying Multi-Layer Perceptrons to Interval-Valued Data},
  volume = {25},
  ISSN = {1573-773X},
  url = {http://dx.doi.org/10.1007/s11063-007-9035-z},
  DOI = {10.1007/s11063-007-9035-z},
  number = {2},
  journal = {Neural Processing Letters},
  publisher = {Springer Science and Business Media LLC},
  author = {Roque,  Antonio Munoz San and Mate,  Carlos and Arroyo,  Javier and Sarabia,  Angel},
  year = {2007},
  month = feb,
  pages = {157-169}
}

@article{Maia2008,
  title = {Forecasting models for interval-valued time series},
  volume = {71},
  ISSN = {0925-2312},
  url = {http://dx.doi.org/10.1016/j.neucom.2008.02.022},
  DOI = {10.1016/j.neucom.2008.02.022},
  number = {16-18},
  journal = {Neurocomputing},
  publisher = {Elsevier BV},
  author = {Maia,  Andre Luis S. and de Carvalho,  Francisco de A.T. and Ludermir,  Teresa B.},
  year = {2008},
  month = oct,
  pages = {3344-3352}
}

@article{Maia2011,
  title = {Holt's exponential smoothing and neural network models for forecasting interval-valued time series},
  volume = {27},
  ISSN = {0169-2070},
  url = {http://dx.doi.org/10.1016/j.ijforecast.2010.02.012},
  DOI = {10.1016/j.ijforecast.2010.02.012},
  number = {3},
  journal = {International Journal of Forecasting},
  publisher = {Elsevier BV},
  author = {Maia, Andre Luis Santiago and de Carvalho,  Francisco de A.T.},
  year = {2011},
  month = jul,
  pages = {740-759}
}

@article{Yang2019,
  title = {Interval-valued data prediction via regularized artificial neural network},
  volume = {331},
  ISSN = {0925-2312},
  url = {http://dx.doi.org/10.1016/j.neucom.2018.11.063},
  DOI = {10.1016/j.neucom.2018.11.063},
  journal = {Neurocomputing},
  publisher = {Elsevier BV},
  author = {Yang,  Zebin and Lin,  Dennis K.J. and Zhang,  Aijun},
  year = {2019},
  month = feb,
  pages = {336-345}
}

@article{Bernardini2021,
  title = {STREAmS: A high-fidelity accelerated solver for direct numerical simulation of compressible turbulent flows},
  volume = {263},
  ISSN = {0010-4655},
  url = {http://dx.doi.org/10.1016/j.cpc.2021.107906},
  DOI = {10.1016/j.cpc.2021.107906},
  journal = {Computer Physics Communications},
  publisher = {Elsevier BV},
  author = {Bernardini,  Matteo and Modesti,  Davide and Salvadore,  Francesco and Pirozzoli,  Sergio},
  year = {2021},
  month = {jun},
  pages = {107906}
}

@article{Huthwaite2014,
  title = {Accelerated finite element elastodynamic simulations using the GPU},
  volume = {257},
  ISSN = {0021-9991},
  url = {http://dx.doi.org/10.1016/j.jcp.2013.10.017},
  DOI = {10.1016/j.jcp.2013.10.017},
  journal = {Journal of Computational Physics},
  publisher = {Elsevier BV},
  author = {Huthwaite,  Peter},
  year = {2014},
  month = {jan},
  pages = {687-707}
}

@article{Lu2021,
  title = {Learning nonlinear operators via DeepONet based on the universal approximation theorem of operators},
  volume = {3},
  ISSN = {2522-5839},
  url = {http://dx.doi.org/10.1038/s42256-021-00302-5},
  DOI = {10.1038/s42256-021-00302-5},
  number = {3},
  journal = {Nature Machine Intelligence},
  publisher = {Springer Science and Business Media LLC},
  author = {Lu,  Lu and Jin,  Pengzhan and Pang,  Guofei and Zhang,  Zhongqiang and Karniadakis,  George Em},
  year = {2021},
  month = {mar},
  pages = {218-229}
}

@misc{Kovachki2021,
  doi = {10.48550/ARXIV.2108.08481},
  url = {https://arxiv.org/abs/2108.08481},
  author = {Kovachki,  Nikola and Li,  Zongyi and Liu,  Burigede and Azizzadenesheli,  Kamyar and Bhattacharya,  Kaushik and Stuart,  Andrew and Anandkumar,  Anima},
  title = {Neural Operator: Learning Maps Between Function Spaces},
  publisher = {arXiv},
  year = {2021},
  copyright = {arXiv.org perpetual,  non-exclusive license}
}

@misc{alkin2025,
      title={Universal Physics Transformers: A Framework For Efficiently Scaling Neural Operators}, 
      author={Benedikt Alkin and Andreas FÃ¼rst and Simon Schmid and Lukas Gruber and Markus Holzleitner and Johannes Brandstetter},
      year={2025},
      eprint={2402.12365},
      archivePrefix={arXiv},
      primaryClass={cs.LG},
      url={https://arxiv.org/abs/2402.12365}, 
}

@misc{sanchezgonzalez2020,
      title={Learning to Simulate Complex Physics with Graph Networks}, 
      author={Alvaro Sanchez-Gonzalez and Jonathan Godwin and Tobias Pfaff and Rex Ying and Jure Leskovec and Peter W. Battaglia},
      year={2020},
      eprint={2002.09405},
      archivePrefix={arXiv},
      primaryClass={cs.LG},
      url={https://arxiv.org/abs/2002.09405}, 
}

@ARTICLE{chen1995,
  author={Tianping Chen and Hong Chen},
  journal={IEEE Transactions on Neural Networks}, 
  title={Universal approximation to nonlinear operators by neural networks with arbitrary activation functions and its application to dynamical systems}, 
  year={1995},
  volume={6},
  number={4},
  pages={911-917},
  doi={10.1109/72.392253}}

@misc{gowal2019,
      title={On the Effectiveness of Interval Bound Propagation for Training Verifiably Robust Models}, 
      author={Sven Gowal and Krishnamurthy Dvijotham and Robert Stanforth and Rudy Bunel and Chongli Qin and Jonathan Uesato and Relja Arandjelovic and Timothy Mann and Pushmeet Kohli},
      year={2019},
      eprint={1810.12715},
      archivePrefix={arXiv},
      primaryClass={cs.LG},
      url={https://arxiv.org/abs/1810.12715}, 
}

@article{Singh2019,
  title = {An abstract domain for certifying neural networks},
  volume = {3},
  ISSN = {2475-1421},
  url = {http://dx.doi.org/10.1145/3290354},
  DOI = {10.1145/3290354},
  number = {POPL},
  journal = {Proceedings of the ACM on Programming Languages},
  publisher = {Association for Computing Machinery (ACM)},
  author = {Singh,  Gagandeep and Gehr,  Timon and P\"{u}schel,  Markus and Vechev,  Martin},
  year = {2019},
  month = jan,
  pages = {1-30}
}

@misc{zhang2018,
      title={Efficient Neural Network Robustness Certification with General Activation Functions}, 
      author={Huan Zhang and Tsui-Wei Weng and Pin-Yu Chen and Cho-Jui Hsieh and Luca Daniel},
      year={2018},
      eprint={1811.00866},
      archivePrefix={arXiv},
      primaryClass={cs.LG},
      url={https://arxiv.org/abs/1811.00866}, 
}

@misc{zhang2019,
      title={Towards Stable and Efficient Training of Verifiably Robust Neural Networks}, 
      author={Huan Zhang and Hongge Chen and Chaowei Xiao and Sven Gowal and Robert Stanforth and Bo Li and Duane Boning and Cho-Jui Hsieh},
      year={2019},
      eprint={1906.06316},
      archivePrefix={arXiv},
      primaryClass={cs.LG},
      url={https://arxiv.org/abs/1906.06316}, 
}

@article{wang2021beta,
  title={{Beta-CROWN}: Efficient bound propagation with per-neuron split constraints for complete and incomplete neural network verification},
  author={Wang, Shiqi and Zhang, Huan and Xu, Kaidi and Lin, Xue and Jana, Suman and Hsieh, Cho-Jui and Kolter, J Zico},
  journal={Advances in Neural Information Processing Systems},
  volume={34},
  year={2021}
}

@misc{xu2020,
      title={Automatic Perturbation Analysis for Scalable Certified Robustness and Beyond}, 
      author={Kaidi Xu and Zhouxing Shi and Huan Zhang and Yihan Wang and Kai-Wei Chang and Minlie Huang and Bhavya Kailkhura and Xue Lin and Cho-Jui Hsieh},
      year={2020},
      eprint={2002.12920},
      archivePrefix={arXiv},
      primaryClass={cs.LG},
      url={https://arxiv.org/abs/2002.12920}, 
}

@article{Ishibuchi1993,
  title = {An architecture of neural networks with interval weights and its application to fuzzy regression analysis},
  volume = {57},
  ISSN = {0165-0114},
  url = {http://dx.doi.org/10.1016/0165-0114(93)90118-2},
  DOI = {10.1016/0165-0114(93)90118-2},
  number = {1},
  journal = {Fuzzy Sets and Systems},
  publisher = {Elsevier BV},
  author = {Ishibuchi,  Hisao and Tanaka,  Hideo and Okada,  Hidehiko},
  year = {1993},
  month = jul,
  pages = {27-39}
}

@article{Oala2021,
  title = {Detecting failure modes in image reconstructions with interval neural network uncertainty},
  volume = {16},
  ISSN = {1861-6429},
  url = {http://dx.doi.org/10.1007/s11548-021-02482-2},
  DOI = {10.1007/s11548-021-02482-2},
  number = {12},
  journal = {International Journal of Computer Assisted Radiology and Surgery},
  publisher = {Springer Science and Business Media LLC},
  author = {Oala,  Luis and Heiss,  Cosmas and Macdonald,  Jan and Marz,  Maximilian and Kutyniok,  Gitta and Samek,  Wojciech},
  year = {2021},
  month = sep,
  pages = {2089-2097}
}

@article{Betancourt2022,
  title = {Interval deep learning for computational mechanics problems under input uncertainty},
  volume = {70},
  ISSN = {0266-8920},
  url = {http://dx.doi.org/10.1016/j.probengmech.2022.103370},
  DOI = {10.1016/j.probengmech.2022.103370},
  journal = {Probabilistic Engineering Mechanics},
  publisher = {Elsevier BV},
  author = {Betancourt,  David and Muhanna,  Rafi L.},
  year = {2022},
  month = oct,
  pages = {103370}
}

@article{Tretiak2023,
  title = {Neural network model for imprecise regression with interval dependent variables},
  volume = {161},
  ISSN = {0893-6080},
  url = {http://dx.doi.org/10.1016/j.neunet.2023.02.005},
  DOI = {10.1016/j.neunet.2023.02.005},
  journal = {Neural Networks},
  publisher = {Elsevier BV},
  author = {Tretiak,  Krasymyr and Schollmeyer,  Georg and Ferson,  Scott},
  year = {2023},
  month = apr,
  pages = {550-564}
}

@inbook{Neumaier1993,
  title = {The Wrapping Effect,  Ellipsoid Arithmetic,  Stability and Confidence Regions},
  ISBN = {9783709169186},
  ISSN = {0344-8029},
  url = {http://dx.doi.org/10.1007/978-3-7091-6918-6_14},
  DOI = {10.1007/978-3-7091-6918-6_14},
  booktitle = {Validation Numerics},
  publisher = {Springer Vienna},
  author = {Neumaier,  A.},
  year = {1993},
  chapter = {0},
  pages = {175-190}
}

@misc{revol2022affineiterationswrappingeffect,
      title={Affine Iterations and Wrapping Effect: Various Approaches}, 
      author={Nathalie Revol},
      year={2022},
      eprint={2201.00513},
      archivePrefix={arXiv},
      primaryClass={math.NA},
      url={https://arxiv.org/abs/2201.00513}, 
}

@article{Wang2025,
  title = {CreINNs: Credal-Set Interval Neural Networks for Uncertainty Estimation in Classification Tasks},
  volume = {185},
  ISSN = {0893-6080},
  url = {http://dx.doi.org/10.1016/j.neunet.2025.107198},
  DOI = {10.1016/j.neunet.2025.107198},
  journal = {Neural Networks},
  publisher = {Elsevier BV},
  author = {Wang,  Kaizheng and Shariatmadar,  Keivan and Manchingal,  Shireen Kudukkil and Cuzzolin,  Fabio and Moens,  David and Hallez,  Hans},
  year = {2025},
  month = may,
  pages = {107198}
}

@article{Varian1975,
author="Varian, H. R.",
title="A Bayesian Approach to Real Estate Assessment",
journal="Studies in Bayesian Econometrics and Statistics in Honor of Leonard J. Savage",
publisher="North-Holland",
year="1975",
URL="https://cir.nii.ac.jp/crid/1572824500518986496"
}

@article{Fu2023,
  title = {Linear-exponential loss incorporated deep learning for imbalanced classification},
  volume = {140},
  ISSN = {0019-0578},
  url = {http://dx.doi.org/10.1016/j.isatra.2023.06.016},
  DOI = {10.1016/j.isatra.2023.06.016},
  journal = {ISA Transactions},
  publisher = {Elsevier BV},
  author = {Fu,  Saiji and Su,  Duo and Li,  Shilin and Sun,  Shiding and Tian,  Yingjie},
  year = {2023},
  month = sep,
  pages = {279-292}
}

@inproceedings{macqueen1967,
  title={Multivariate observations},
  author={MacQueen, J},
  booktitle={Proceedings ofthe 5th Berkeley Symposium on Mathematical Statisticsand Probability},
  volume={1},
  pages={281--297},
  year={1967}
}

@article{lloyd1982least,
  title={Least squares quantization in PCM},
  author={Lloyd, Stuart},
  journal={IEEE transactions on information theory},
  volume={28},
  number={2},
  pages={129--137},
  year={1982},
  publisher={IEEE}
}

@article{tripathy2018deep,
  title={Deep UQ: Learning deep neural network surrogate models for high dimensional uncertainty quantification},
  author={Tripathy, Rohit K and Bilionis, Ilias},
  journal={Journal of computational physics},
  volume={375},
  pages={565--588},
  year={2018},
  publisher={Elsevier}
}

@article{sun2019review,
  title={A review of the artificial neural network surrogate modeling in aerodynamic design},
  author={Sun, Gang and Wang, Shuyue},
  journal={Proceedings of the Institution of Mechanical Engineers, Part G: Journal of Aerospace Engineering},
  volume={233},
  number={16},
  pages={5863--5872},
  year={2019},
  publisher={SAGE Publications Sage UK: London, England}
}

@BOOK{Rasmussen2005-at,
  title     = "Gaussian processes for machine learning",
  author    = "Rasmussen, Carl Edward and Williams, Christopher K I",
  publisher = "MIT Press",
  series    = "Adaptive Computation and Machine Learning series",
  month     =  nov,
  year      =  2005,
  address   = "London, England",
  language  = "en"
}

@article{Ghanem1990,
  doi = {10.1115/1.2888303},
  url = {https://doi.org/10.1115/1.2888303},
  year = {1990},
  month = mar,
  publisher = {{ASME} International},
  volume = {57},
  number = {1},
  pages = {197--202},
  author = {Roger Ghanem and P. D. Spanos},
  title = {Polynomial Chaos in Stochastic Finite Elements},
  journal = {Journal of Applied Mechanics}
}

@article{Abdi2023,
  title = {Propagating input uncertainties into parameter uncertainties and model prediction uncertainties: A review},
  volume = {102},
  ISSN = {1939-019X},
  url = {http://dx.doi.org/10.1002/cjce.25015},
  DOI = {10.1002/cjce.25015},
  number = {1},
  journal = {The Canadian Journal of Chemical Engineering},
  publisher = {Wiley},
  author = {Abdi,  Kaveh and Celse,  Benoit and McAuley,  Kim},
  year = {2023},
  month = jun,
  pages = {254-273}
}

@article{Lu2022,
  title = {A comprehensive and fair comparison of two neural operators (with practical extensions) based on FAIR data},
  volume = {393},
  ISSN = {0045-7825},
  url = {http://dx.doi.org/10.1016/j.cma.2022.114778},
  DOI = {10.1016/j.cma.2022.114778},
  journal = {Computer Methods in Applied Mechanics and Engineering},
  publisher = {Elsevier BV},
  author = {Lu,  Lu and Meng,  Xuhui and Cai,  Shengze and Mao,  Zhiping and Goswami,  Somdatta and Zhang,  Zhongqiang and Karniadakis,  George Em},
  year = {2022},
  month = {apr},
  pages = {114778}
}

@misc{Dai2024,
  doi = {10.48550/ARXIV.2405.02461},
  url = {https://arxiv.org/abs/2405.02461},
  author = {Dai,  Jin and Adhikari,  Santosh and Wen,  Mingjian},
  title = {Uncertainty Quantification and Propagation in Atomistic Machine Learning},
  publisher = {arXiv},
  year = {2024},
  copyright = {arXiv.org perpetual,  non-exclusive license}
}

@article{Dick2013,
  title = {High-dimensional integration: The quasi-Monte Carlo way},
  volume = {22},
  ISSN = {1474-0508},
  url = {http://dx.doi.org/10.1017/S0962492913000044},
  DOI = {10.1017/s0962492913000044},
  journal = {Acta Numerica},
  publisher = {Cambridge University Press (CUP)},
  author = {Dick,  Josef and Kuo,  Frances Y. and Sloan,  Ian H.},
  year = {2013},
  month = apr,
  pages = {133-288}
}

@book{Lemieux2009,
  ISBN = {9780387781655},
  ISSN = {0172-7397},
  title = {Monte Carlo and Quasi-Monte Carlo Sampling},
  author    = {Lemieux, Christine},
  url = {http://dx.doi.org/10.1007/978-0-387-78165-5},
  DOI = {10.1007/978-0-387-78165-5},
  journal = {Springer Series in Statistics},
  publisher = {Springer New York},
  year = {2009}
}

@article{Callens2022,
  title = {MULTILEVEL QUASI-MONTE CARLO FOR INTERVAL ANALYSIS},
  volume = {12},
  ISSN = {2152-5080},
  url = {http://dx.doi.org/10.1615/Int.J.UncertaintyQuantification.2022039245},
  DOI = {10.1615/int.j.uncertaintyquantification.2022039245},
  number = {4},
  journal = {International Journal for Uncertainty Quantification},
  publisher = {Begell House},
  author = {Callens,  Robin R.P. and Faess,  Matthias G.R. and Moens,  David},
  year = {2022},
  pages = {1-19}
}

@article{Cicirello2022,
  title = {Machine learning based optimization for interval uncertainty propagation},
  volume = {170},
  ISSN = {0888-3270},
  url = {http://dx.doi.org/10.1016/j.ymssp.2021.108619},
  DOI = {10.1016/j.ymssp.2021.108619},
  journal = {Mechanical Systems and Signal Processing},
  publisher = {Elsevier BV},
  author = {Cicirello,  Alice and Giunta,  Filippo},
  year = {2022},
  month = may,
  pages = {108619}
}

@article{White2019,
  title = {Multiscale topology optimization using neural network surrogate models},
  volume = {346},
  ISSN = {0045-7825},
  url = {http://dx.doi.org/10.1016/j.cma.2018.09.007},
  DOI = {10.1016/j.cma.2018.09.007},
  journal = {Computer Methods in Applied Mechanics and Engineering},
  publisher = {Elsevier BV},
  author = {White,  Daniel A. and Arrighi,  William J. and Kudo,  Jun and Watts,  Seth E.},
  year = {2019},
  month = apr,
  pages = {1118-1135}
}

@article{Zhang2021,
  title = {Multi-fidelity deep neural network surrogate model for aerodynamic shape optimization},
  volume = {373},
  ISSN = {0045-7825},
  url = {http://dx.doi.org/10.1016/j.cma.2020.113485},
  DOI = {10.1016/j.cma.2020.113485},
  journal = {Computer Methods in Applied Mechanics and Engineering},
  publisher = {Elsevier BV},
  author = {Zhang,  Xinshuai and Xie,  Fangfang and Ji,  Tingwei and Zhu,  Zaoxu and Zheng,  Yao},
  year = {2021},
  month = jan,
  pages = {113485}
}

@article{Sudret2008,
  title = {Global sensitivity analysis using polynomial chaos expansions},
  volume = {93},
  ISSN = {0951-8320},
  url = {http://dx.doi.org/10.1016/j.ress.2007.04.002},
  DOI = {10.1016/j.ress.2007.04.002},
  number = {7},
  journal = {Reliability Engineering and System Safety},
  publisher = {Elsevier BV},
  author = {Sudret,  Bruno},
  year = {2008},
  month = jul,
  pages = {964-979}
}

@article{Bilionis2012,
  title = {Multi-output local Gaussian process regression: Applications to uncertainty quantification},
  volume = {231},
  ISSN = {0021-9991},
  url = {http://dx.doi.org/10.1016/j.jcp.2012.04.047},
  DOI = {10.1016/j.jcp.2012.04.047},
  number = {17},
  journal = {Journal of Computational Physics},
  publisher = {Elsevier BV},
  author = {Bilionis,  Ilias and Zabaras,  Nicholas},
  year = {2012},
  month = jul,
  pages = {5718-5746}
}

@article{Zuhal2023,
  title = {Performance assessment of Kriging with partial least squares for high-dimensional uncertainty and sensitivity analysis},
  volume = {66},
  ISSN = {1615-1488},
  url = {http://dx.doi.org/10.1007/s00158-023-03547-3},
  DOI = {10.1007/s00158-023-03547-3},
  number = {5},
  journal = {Structural and Multidisciplinary Optimization},
  publisher = {Springer Science and Business Media LLC},
  author = {Zuhal,  Lavi Rizki and Faza,  Ghifari Adam and Palar,  Pramudita Satria and Liem,  Rhea Patricia},
  year = {2023},
  month = apr 
}

@article{Liu2026,
  title = {A novel study on hybrid physics-data-driven reduced-order modeling for aerodynamic load inversion under structural field uncertainties},
  volume = {448},
  ISSN = {0045-7825},
  url = {http://dx.doi.org/10.1016/j.cma.2025.118504},
  DOI = {10.1016/j.cma.2025.118504},
  journal = {Computer Methods in Applied Mechanics and Engineering},
  publisher = {Elsevier BV},
  author = {Liu,  Yaru and Wang,  Lei and Zhou,  Xuan and Li,  Zeshang and Wang,  Yuewu},
  year = {2026},
  month = jan,
  pages = {118504}
}

@article{Ma2025,
  title = {Data-Driven interval robust optimization method of VPP Bidding strategy in spot market under multiple uncertainties},
  volume = {384},
  ISSN = {0306-2619},
  url = {http://dx.doi.org/10.1016/j.apenergy.2025.125366},
  DOI = {10.1016/j.apenergy.2025.125366},
  journal = {Applied Energy},
  publisher = {Elsevier BV},
  author = {Ma,  Ying and Li,  Zhen and Liu,  Ruyi and Liu,  Bin and Yu,  Samson S. and Liao,  Xiaozhong and Shi,  Peng},
  year = {2025},
  month = apr,
  pages = {125366}
}

@article{Zhao2025,
  title = {Double-scale time-dependent reliable topology optimization based on the first-passage failure and interval process theories},
  volume = {443},
  ISSN = {0045-7825},
  url = {http://dx.doi.org/10.1016/j.cma.2025.118088},
  DOI = {10.1016/j.cma.2025.118088},
  journal = {Computer Methods in Applied Mechanics and Engineering},
  publisher = {Elsevier BV},
  author = {Zhao,  Xingyu and Wang,  Lei},
  year = {2025},
  month = aug,
  pages = {118088}
}

@article{Zhang2025,
  title = {Interval-updated inverse identification framework for transient coupling of loads and heat fluxes via force-thermal modal decomposition and adaptive surrogate modeling},
  volume = {166},
  ISSN = {0735-1933},
  url = {http://dx.doi.org/10.1016/j.icheatmasstransfer.2025.109223},
  DOI = {10.1016/j.icheatmasstransfer.2025.109223},
  journal = {International Communications in Heat and Mass Transfer},
  publisher = {Elsevier BV},
  author = {Zhang,  Haoyu and Wang,  Lei},
  year = {2025},
  month = aug,
  pages = {109223}
}

@article{Koenker1978,
  title = {Regression Quantiles},
  volume = {46},
  ISSN = {0012-9682},
  url = {http://dx.doi.org/10.2307/1913643},
  DOI = {10.2307/1913643},
  number = {1},
  journal = {Econometrica},
  publisher = {JSTOR},
  author = {Koenker,  Roger and Bassett,  Gilbert},
  year = {1978},
  month = jan,
  pages = {33}
}

@article{Steinwart2011,
   title={Estimating conditional quantiles with the help of the pinball loss},
   volume={17},
   ISSN={1350-7265},
   url={http://dx.doi.org/10.3150/10-BEJ267},
   DOI={10.3150/10-bej267},
   number={1},
   journal={Bernoulli},
   publisher={Bernoulli Society for Mathematical Statistics and Probability},
   author={Steinwart, Ingo and Christmann, Andreas},
   year={2011},
   month=feb }

@misc{Pouplin2025,
      title={Relaxed Quantile Regression: Prediction Intervals for Asymmetric Noise}, 
      author={Thomas Pouplin and Alan Jeffares and Nabeel Seedat and Mihaela van der Schaar},
      year={2025},
      eprint={2406.03258},
      archivePrefix={arXiv},
      primaryClass={stat.ML},
      url={https://arxiv.org/abs/2406.03258}, 
}

@BOOK{Gammerman2005CP,
  title     = "Algorithmic learning in a random world",
  author    = "Gammerman, Alex and Shafer, Glenn and Vovk, Vladimir",
  publisher = "Springer Science+Business Media",
  month     =  jan,
  year      =  2005,
  language  = "en"
}

@inbook{Papadopoulos2002CP,
  title = {Inductive Confidence Machines for Regression},
  ISBN = {9783540367550},
  ISSN = {0302-9743},
  url = {http://dx.doi.org/10.1007/3-540-36755-1_29},
  DOI = {10.1007/3-540-36755-1_29},
  booktitle = {Machine Learning: ECML 2002},
  publisher = {Springer Berlin Heidelberg},
  author = {Papadopoulos,  Harris and Proedrou,  Kostas and Vovk,  Volodya and Gammerman,  Alex},
  year = {2002},
  chapter={1},
  pages = {345-356}
}







\end{document}